%% file: main.tex
\documentclass[10pt,twocolumn,letterpaper]{article}

\usepackage{cvpr} % To force page numbers, e.g. for an arXiv version

\usepackage{array}
\usepackage{colortbl}
\usepackage{siunitx}
\usepackage{multirow}
\usepackage{graphicx}
\usepackage{amsmath}
\usepackage{rotating}
\usepackage{makecell}
\usepackage{float}

\definecolor{cvprblue}{rgb}{0.21,0.49,0.74}
\usepackage[pagebackref,breaklinks,colorlinks,allcolors=cvprblue]{hyperref}

\def\paperID{} % *** Enter the Paper ID here
\def\confName{CVPR}
\def\confYear{2025}

\title{Spatial-Temporal Graph Diffusion Policy with Kinematic Modeling for Bimanual Robotic Manipulation}

\author{Qi Lv$^{1\text{,}2,\text{3}}$ \quad
Hao Li$^{\text{1,3}}$ \quad
Xiang Deng$^{\text{1}\dagger}$ \quad
Rui Shao$^{\text{1}}$ \quad 
Yinchuan Li$^{\text{3}}$ \quad 
Jianye Hao$^{\text{3}}$\\
Longxiang Gao$^{\text{4}}$ \quad
Michael Yu Wang$^{\text{2}}$ \quad
Liqiang Nie$^{\text{1}\dagger}$
\\
[1mm]
$^{1}$Harbin Institute of Technology (Shenzhen) \quad
$^{2}$Greate Bay University \\
$^{3}$Huawei Noah's Ark Lab \quad
$^{4}$Shandong Computer Science Center \\
[1mm]
\texttt{lvqi@stu.hit.edu.cn,} \texttt{dengxiang@hit.edu.cn}
}

\begin{document}
\maketitle

{
\let\thefootnote \relax \footnote{
$^*$This work was done when Qi Lv and Hao Li were intern at Huawei Noah's Ark Lab. $^\dagger$Corresponding author}
}

\begin{abstract}
Despite the significant success of imitation learning in robotic manipulation, its application to bimanual tasks remains highly challenging. 
Existing approaches mainly learn a policy to predict a distant next-best end-effector pose (NBP) and then compute the corresponding joint rotation angles for motion using inverse kinematics.
However, they suffer from two important issues: \textbf{(1) rarely considering the physical robotic structure}, which may cause self-collisions or interferences, and \textbf{(2) overlooking the kinematics constraint}, which may result in the predicted poses not conforming to the actual limitations of the robot joints.
In this paper, we propose \textbf{K}inematics enhanced \textbf{S}patial-\textbf{T}empor\textbf{A}l g\textbf{R}aph \textbf{Diffuser} (\textbf{KStar Diffuser}).
Specifically, \textbf{(1)} to incorporate the physical robot structure information into action prediction, KStar Diffuser maintains a dynamic spatial-temporal graph according to the physical bimanual joint motions at continuous timesteps. 
This dynamic graph serves as the robot-structure condition for denoising the actions;
\textbf{(2)} to make the NBP learning objective consistent with kinematics, we introduce the differentiable kinematics to provide the reference for optimizing KStar Diffuser.
This module regularizes the policy to predict more reliable and kinematics-aware next end-effector poses.
Experimental results show that our method effectively leverages the physical structural information and generates kinematics-aware actions in both simulation and real-world.%\footnote{Code will be released upon the acceptance.}.
\end{abstract}

\section{Introduction}
\label{sec:intro}
Bimanual manipulation~\cite{smith2012dual, sarkar1997dynamic, paljug1994control, nakamura1989dynamics} represents a fundamental capability for robotic systems to perform complex tasks requiring two-arm coordination. 
While imitation learning has demonstrated remarkable success in single-arm manipulation~\cite{brohan2023rt, brohan2022rt, mu2024embodiedgpt, NEURIPS2024_288b63aa, }, its extension to bimanual scenarios faces unique challenges as robots need to coordinate dual-arm movements while conforming to physical constraints. 
These challenges significantly impact the reliability and feasibility of predicted actions in real-world applications.

\begin{figure}[tbp]
    \centering
    \includegraphics[width=0.85\linewidth]{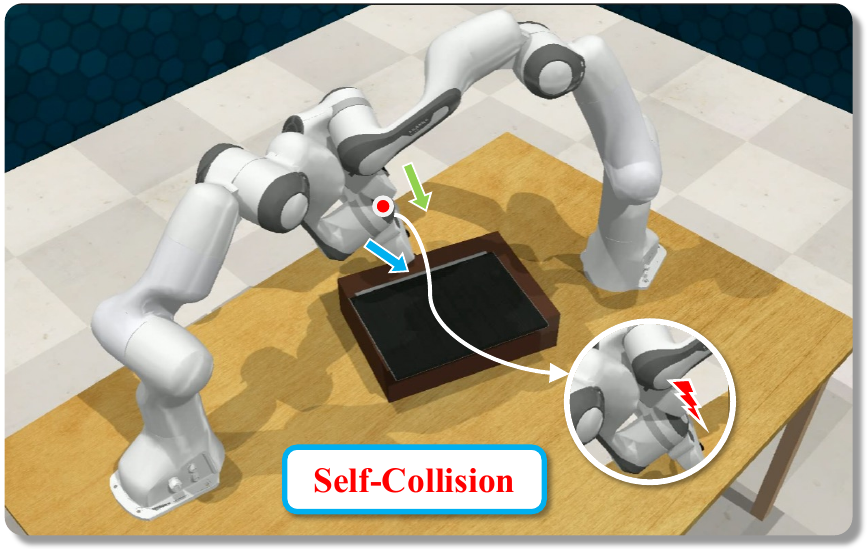}
    \caption{The self-collision problem in bimanual manipulation tasks due to overlooking the robotic structure.}
    \label{fig:abs}
    \vspace{-3.5mm}
\end{figure}

A straightforward approach~\cite{zhao2023learning, chi2023diffusion, ze20243d, fu2024mobile, chi2024universal, ze2024generalizable} would be to learn the entire motion trajectory by predicting joint positions at consecutive timesteps. 
However, this poses significant challenges as trajectories are typically long and difficult to learn. 
Therefore, mainstream approaches~\cite{goyal2023rvt, grotz2024peract2, shridhar2023perceiver, ma2024hierarchical} typically adopt a two-stage pipeline: first predicting the next-best end-effector pose (NBP), then computing joint rotations through inverse kinematics~\cite{lee1982robot, hunt1983structural}.
However, the dominant NBP-based approaches, though simplify the learning process, often lead to unreliable motion generation. 
Concretely, the predicted poses may violate physical structure constraints~\cite{lei2020real, shin1989collision}, causing potential arm collisions, or exceeding joint limits due to inadequate consideration of kinematic feasibility, as Figure~\ref{fig:abs} illustrated.

In real-world bimanual manipulation tasks, existing approaches~\cite{grotz2024peract2, chi2023diffusion, goyal2023rvt, shridhar2023perceiver, zhang2024sam} exhibit significant limitations in motion execution despite their promising performance in trajectory prediction. 
Empirical observations indicate that while individual end-effector poses are feasible in isolation, their concurrent execution frequently results in inter-arm collisions~\cite{shin1989collision} and kinematically infeasible configurations~\cite{ma2024hierarchical, ortenzi2018vision}.
The fundamental challenge lies in the optimization of end-effector poses exclusively in Cartesian space, which introduces a substantial discrepancy between motion prediction and physical constraints. 
Existing approaches~\cite{grotz2024peract2, chi2023diffusion, chi2024universal} incorporate proprioception information in a simplistic manner, (\eg, gripper rotation angles~\cite{shridhar2023perceiver, grotz2024peract2} and end-effector poses~\cite{wang2024scaling}), but disregarding the robot's intrinsic structural properties \eg, the kinematic chain and joint configurations during the prediction phase. 
This oversimplified representation fails to capture the crucial spatial correlations between robotic arms, joint links, and their potential interactions.
Consequently, the predicted dual-arm trajectories may satisfy task objectives in Cartesian space while violating crucial spatial constraints during execution.

Moreover, the conventional paradigm~\cite{kim2024openvla, grotz2024peract2, shridhar2023perceiver, team2024octo} of treating inverse kinematics as a post-processing procedure introduces additional complexities. 
The disjunction between kinematic constraints and the pose prediction learning objective often yields solutions that exhibit apparent smoothness in Cartesian space while manifesting discontinuous or mechanically infeasible trajectories in the joint space. 
This limitation becomes particularly pronounced in configurations approaching kinematic singularities or joint limits, where the mapping between Cartesian and joint spaces becomes ill-conditioned. 
These inherent limitations necessitate the integration of both structural and kinematic constraints into the policy learning framework.

To address the above issues, we propose \textbf{K}inematics enhanced \textbf{S}patial-\textbf{T}empor\textbf{A}l g\textbf{R}aph \textbf{Diffuser} (\textbf{KStar Diffuser}), a novel framework that explicitly incorporates both robot structures and kinematics into the bimanual motion generation process. 
Our key insight is that the physical structure and kinematic properties of the robot should guide the learning process of pose prediction, rather than be treated as independent post-processing constraints.
Specifically, \textbf{(1) to incorporate structural awareness}, we construct a dynamic spatial-temporal graph from the robot's URDF specifications, where nodes represent joint properties and edges capture both spatial relations and temporal dependencies. 
This graph structure is encoded via Graph Convolutional Network (GCN)~\cite{kipf2016semi} to provide explicit physical constraints for the diffusion process;
\textbf{(2) for kinematic feasibility}, we regularize the NBP learning objective by incorporating joint-space prediction, where the predicted joint positions are mapped to reference end-effector poses through differentiable forward kinematics.
These kinematically-feasible poses then serve as conditions to guide the diffusion process, ensuring that the generated motions satisfy both structural and kinematic constraints.
Our main contributions are as follows:
\begin{itemize}
    \item Different from existing approaches that optimize end-effector poses solely in Cartesian space, we propose a novel spatial-temporal robot graph that explicitly models the robot physical configuration to guide the generative action denoising procedure.
    \item We introduce a kinematics regularizer that augments the NBP learning objective by introducing joint-space supervision. This regularizer leverages forward kinematics to provide kinematically-feasible reference poses, effectively guiding the diffusion process to conform to kinematic constraints.
    \item Extensive experiments show that our proposed KStar Diffuser is superior in both simulation and real-world scenarios, surpassing baselines more than 10\% in success rate.
\end{itemize}
\vspace{-1mm}
\section{Related Work}
\label{sec:formatting}
\vspace{-1mm}
\paragraph{Diffusion Models in Bimanual Robotic Controls.}
Diffusion models such as Denoising Diffusion Probabilistic Models (DDPM)~\cite{ho2020denoising} have achieved great success in the fields of image generation~\cite{rombach2022high, ramesh2022hierarchical, podell2023sdxl} and video generation~\cite{peebles2023scalable, wang2024prolificdreamer, zheng1open}. 
Thus, recent work~\cite{chen2023polarnet, zhu2024scaling, wu2024discrete, vosylius2024render, ren2024diffusion} has been devoted to applying its powerful generation capabilities to action generation for robotic manipulation tasks in a imitation learning mode.
With the 2D images~\cite{chi2023diffusion, vosylius2024render, ren2024diffusion, ma2024hierarchical, wang2024sparse, pmlr-v235-zhang24aj, 10239469, li2023mask, li2023fine, li2024genview} or 3D point cloud observations~\cite{chen2023polarnet, ze20243d, ke20243d, ze2024generalizable, huang20243d}, the policy is trained to output the action sequence of joint positions or end-effector poses by iterative denoising process.

Bimanual manipulation tasks, which resemble human-like actions more closely, have garnered increasing attention~\cite{zhao2023learning, gao2024bi, grotz2024peract2, liu2024voxact, chi2024universal}. 
However, existing research primarily extends single-arm manipulation methods to dual-arm tasks~\cite{zhao2023learning, chi2024universal}, without considering about the distinct challenges specific to bimanual settings. 
The relative independence of the two arms makes it essential to consider self-collision avoidance~\cite{10000179, bai2021dual} when predicting actions.
In practice, the robot’s structure significantly influences its motion since the joint types determine movement directions and joint angles constrain the range of motion. 
Although some studies~\cite{goyal2023rvt, shridhar2023perceiver, wang2024scaling} integrate proprioception in robotic control, they often use simply low-dimensional vectors to represent body information, lacking depth in structural exploration and overlooking critical spatial relationships between dual arms.
Therefore, we study the physical structure information of the dual-arm robot system, and improve both the accuracy and adaptability of robotic movements.

\vspace{-4mm}
\paragraph{Robotic Kinematics Modeling.}
Robotic kinematics modeling~\cite{ma2024hierarchical, xia2024kinematic, zanchettin2011kinematic, aristidou2018inverse, grassmann2018learning} is a classic problem in the robotic control~\cite{mason2001mechanics, welman1993inverse, hayat2015robot, bestick2015personalized} where inverse kinematics (IK) presents a fundamental challenge, \ie, deriving the joint configuration given the end-effector pose.
The inherent non-uniqueness of IK solutions as a significant obstacle for learning algorithms~\cite{jordan2013forward, d2001learning}, often produce inaccurate models by averaging over nonconvex feasible sets.
Recent advancements integrate neural solutions~\cite{li2021hybrik, 8211457, Li_2021_CVPR} to enhance the adaptability and precision of kinematic models. 
Bócsi \etal.~\cite{bocsi2011learning} leveraged support vector machines to parameterize quadratic programs, aligning solutions with IK in specific workspace regions. 
Current data-driven imitation learning approaches ~\cite{zhao2023learning, goyal2023rvt, grotz2024peract2, chi2024universal, fu2024mobile} primarily rely on probabilistic modeling of action distributions to predict subsequent actions, but lack the reliability guarantees from the aspect of kinematics.
In this paper, we construct a spatial-temporal graph to learn the robot physical representation, and provide a kinematics-awared latent end-effect embedding for diffusion policy as guidance.
In this way, o`ur approach enhances the robot's ability to perform tasks with superious precision and adaptability.

\section{Method}\label{sec:method}
\subsection{Preliminary}
\paragraph{Diffusion Policy.}
Chi \etal~\cite{chi2023diffusion} propose the diffusion policy which represents a robot's visuomotor policy as a conditional denoising diffusion process.
It learns a model distribution $\pi_\phi(\boldsymbol{a}_0|\boldsymbol{o})$ conditioned on observation $\boldsymbol{o}$ to approximate joint distribution $q(\boldsymbol{a}_0)$.
The whole procedure consists of a \textit{forward process} and a \textit{reverse process}.

(1) Forward Process:
For the Markov chain with Gaussian transitions parameterized, the policy $\pi_\phi(\boldsymbol{a}_0|\boldsymbol{o})$ learns a fixed inference procedure $q(\boldsymbol{a}_k|\boldsymbol{a}_0)$:
\begin{equation}
    \begin{split}
        q(\boldsymbol{a}_k|\boldsymbol{a}_0)= \int q(\boldsymbol{a}_{1:k}|\boldsymbol{a}_0)\mathrm{d}\boldsymbol{a}_{1:(k-1)}\\=\mathcal{N}(\boldsymbol{a}_k;\sqrt{\overline{\alpha}_k}\boldsymbol{a}_0,(1-\overline{\alpha}_k)\boldsymbol{I})
    \end{split}
\end{equation}
Thus, $a_k$ can be expressed as a linear combination of $\boldsymbol{a}_0$ and a noise variable $\epsilon$:
\begin{equation}
    \boldsymbol{a}_k=\sqrt{\overline{\alpha}_k}\boldsymbol{a}_0+\sqrt{1-\overline{\alpha}_k}\epsilon
\end{equation}
The training loss is obtained:
\begin{equation}
    \mathcal{L}=\mathbb{E}_{\boldsymbol{a}_0\sim q(\boldsymbol{a}_0),\epsilon_{k}\sim\mathcal{N}(\boldsymbol{0},\boldsymbol{I}),k}\left[||\epsilon_k, \pi_\phi(\boldsymbol{a}_0+\epsilon_k, \boldsymbol{o}, k)||^2\right]
\end{equation}
where $\epsilon_k$ is the random noise sampling at $k$ iteration.

(2) Reverse Process:
Starting from a sample $\boldsymbol{a}_{K}\sim\mathcal{N}(\boldsymbol{0}, \boldsymbol{I})$, the reverse steps are:
\begin{equation}
    \boldsymbol{a}_{k-1}=\sqrt{\overline{\alpha}_{k-1}}\,\pi_{\phi}(a_k, k, o)+\sqrt{1-\overline{\alpha}_{k-1}}\epsilon.
\end{equation}
By the iterative denoising process, the policy generates $\boldsymbol{a}_0$ as its next action.
The continuous predicted action sequence forms the complete action trajectory.

\vspace{-4mm}
\paragraph{Robotic Kinematics.}
Robotics kinematics describes the relationship between the robot joints and its end-effector. 
It can be divided into \textit{forward kinematics} and \textit{inverse kinematics} for controlling robot movements. 

(1) Forward Kinematics (FK): Given a specific configuration of joint angles, i.e., $\theta=[\theta_1, \theta_2, \ldots, \theta_n]\in \Theta$, it aims to construct a function $f_\mathrm{FK}: \mathbb{R}^n\rightarrow SE(3)$, computing the end-effector position and orientation that are represented as a pose matrix $\mathbf{T}\in SE(3)$.
\vspace{-1mm}
\begin{equation}
    FK(\theta) = \mathbf{T}\in\mathcal{T},\ 
    \mathrm{where}\ \  \mathbf{T}=\prod\nolimits_{i=1}^{n} \mathbf{T}_i(\theta_i),
\end{equation}
where $\mathbf{T}_i(\theta_i)$ is the homogeneous transformation matrix for joint $i$, and $\mathcal{T}$ means the task space.
$\Theta\in\mathbb{R}^n$ denotes the configuration space, while the configuration refers to the specific arrangement of the robotic joints in a given pose.

(2) Inverse Kinematics (IK): In contrast, IK aims to find the joint configurations which achieve a desired end-effector pose.
It is defined as a mapping $f_\mathrm{IK}: SE(3) \rightarrow \mathbb{R}^n$ from the task space $\mathcal{T}$ to the set of possible configurations $\Theta$ that satisfy the given pose matrix $\mathbf{T}$:
\begin{equation}
    IK(\mathcal{T}) = \{\theta\in \Theta | FK(\theta) = \mathbf{T}\}.\label{eq:ik}
\end{equation}
It is noted that this mapping is often non-unique, especially for redundant manipulators with more than 6 DoF.

\begin{figure*}[ht]
    \centering
    \includegraphics[width=\linewidth]{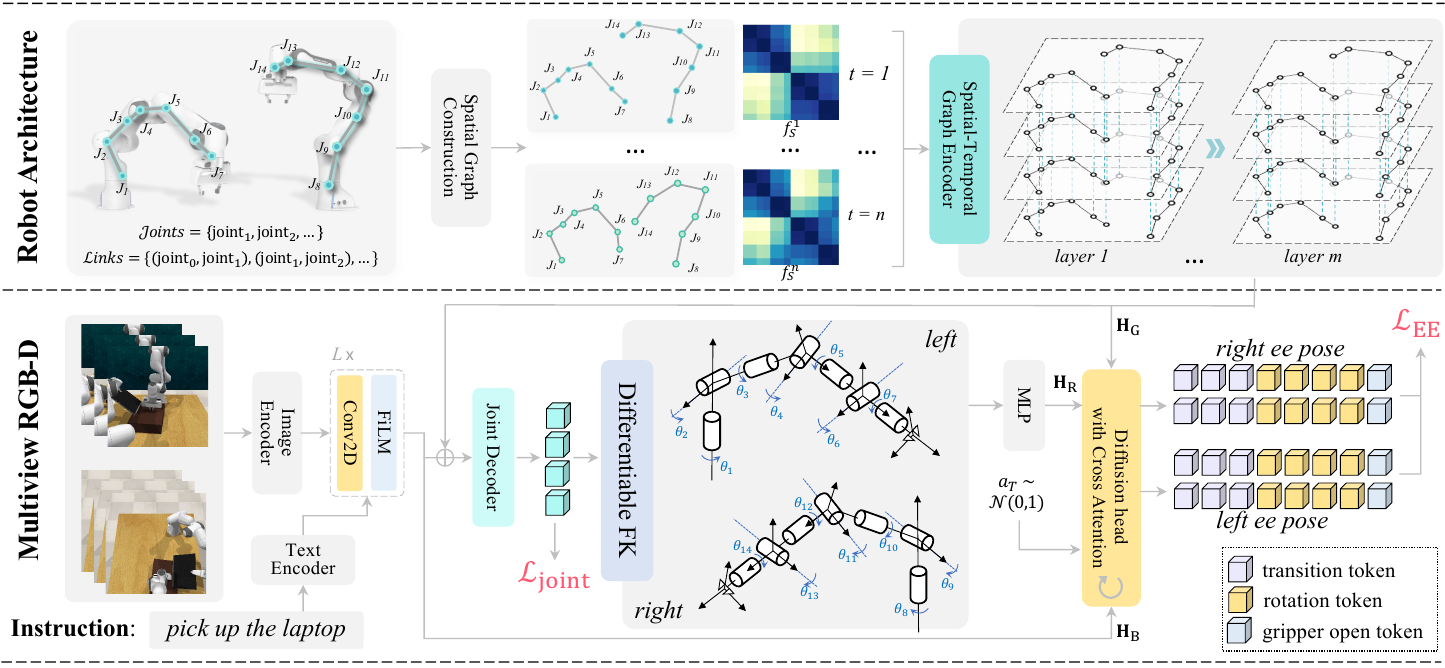}
    \caption{\textbf{Overview of KStar Diffuser}. The top part presents the spatial-temporal graph which is constructed according to the robot architecture. 
    The bottom part shows our backbone and the proposed kinematics regularizer.
    For the backbone, it extracts the multimodal information which consists of multiview RGB-D observations and language instruction, and then generates bimanual 6D end-effector poses.
    The kinematics regularizer enhances pose learning by incorporating joint-level predictions, which are mapped to reference end-effector poses through differentiable forward kinematics (FK).}
    \vspace{-2.5mm}
    \label{fig:framwork}
\end{figure*}

\subsection{Task Definition}
Given the dataset $D$ with language instructions $\boldsymbol{l}$ and the RGB-D observations $\boldsymbol{o}$, we aim to learn a policy $\pi_\phi(\boldsymbol{a}|\boldsymbol{o},\boldsymbol{l})$ which predicts actions $\boldsymbol{a}$.
Here, $\boldsymbol{a}$ is composed of a trajectory $\boldsymbol{a}^\mathrm{joint}=\{\boldsymbol{a}_0, \boldsymbol{a}_1, \ldots, \boldsymbol{a}_T\}$ and gripper opening or closing action $\boldsymbol{a}^\mathrm{gripper}$, where $T$ denotes the trajectory length and $\boldsymbol{a}_i\in\mathbb{R}^n$.
$n$ represents the number of robot joints.
In the bimanual manipulation, $n$ is usually to be 12 or 14 as each robot arm has 6 or 7-DoF.
Referring to prior works~\cite{grotz2024peract2, shridhar2023perceiver, goyal2023rvt, zhang2024sam}, it is inefficient to train the policy on all trajectory points.
Thus, a keyframe discovery method is used to extract a set of $K_\xi$ keyframe indices $\{k_i\}_{i=0}^{K_\xi}$ and the prediction actions are the key set of end-effector poses $\boldsymbol{a}^\mathrm{EE}$.

\subsection{KStar Diffuser}
\subsubsection{Overview}
Mainstream methods~\cite{ke20243d, zhao2023learning, grotz2024peract2} train a policy to predict actions, but take little consideration about the mechanical robot structure which determines its motion.
We thus propose a spatial-temporal graph to model both the static physical structure and dynamic history movement information. 
In addition, to reduce kinematically infeasible predictions for the end-effector pose, we introduce a differentiable kinematics module that provides kinematics-aware references to the policy network.
The overview of our proposed Kinematics enhanced Spatial-TemporAl gRaph Diffuser (KStar Diffuser) is shown in Figure~\ref{fig:framwork}.
\subsubsection{Backbone}
Given the language instruction $l$ and multiview RGB-D observation images $o$, we first adopt the Transformer-based encoders to extract their features $\mathbf{E}_\mathrm{I}$ and $\mathbf{E}_\mathrm{T}$, respectively.
Then the features are fused with the $l_\mathrm{FiLM}$-layers FiLM block~\cite{perez2018film} to obtain the hidden states $\mathbf{H}_\mathrm{B}$.
Each layer is combined with an up-sampling 2D convolution layer:
\begin{equation}
    \mathbf{H}_\mathrm{B}^{i}=\operatorname{Conv2D}(\mathbf{H}_\mathrm{B}^i),
\end{equation}
\begin{equation}
    \mathbf{H}_\mathrm{B}^{i+1}=\operatorname{FiLM}^{i}(\mathbf{H}_\mathrm{B}^{i}, \mathbf{E}_\mathrm{T}),
\end{equation}
where $i\in [0, l_\mathrm{FiLM})$ and the $\mathbf{H}_\mathrm{B}^0$ is initialized with $\mathbf{E}_\mathrm{I}$.

Our backbone $\pi_\phi$ uses the last hidden states $\mathbf{H}_\mathrm{B}$ as the condition to guide a diffusion head to denoise and generate the bimanual end-effector pose.
It is noted that we attach the $n$ historical observation images to provide more information for capturing the motion tendency.
Following Chi~\etal~\cite{chi2023diffusion}, we let the policy predict next $m$ actions during training to alleviate multimodal problems.
Here, we set both $n$ and $m$ to 2.
The action prediction is as follows:
\begin{equation}
    \boldsymbol{a}_{k-1}^{\mathrm{EE}} \sim \pi_\phi(\boldsymbol{a}_{k-1}^{\mathrm{EE}}|\boldsymbol{a}_{k}^{\mathrm{EE}}, \mathbf{H}_\mathrm{B}), \label{eq:original_diffusion_action}
\end{equation}
\begin{equation}
    \begin{split}
        &\pi_\phi(\boldsymbol{a}_{k-1}^{\mathrm{EE}}|\boldsymbol{a}_{k}^{\mathrm{EE}}, \mathbf{H}_\mathrm{B})\\
        &=\mathcal{N}(\boldsymbol{a}_{k-1}^{\mathrm{EE}};\sqrt{\overline{\alpha}_{k-1}}\ \pi_\phi(\boldsymbol{a}_k^{\mathrm{EE}}, \mathbf{H}_\mathrm{B});(1-\overline{\alpha}_{k-1})I).\label{eq:original_diffusion_action_prob}
    \end{split}
\end{equation}
The learning objective is:
\begin{equation}
\mathcal{L}_\mathrm{EE}=\mathbb{E}_{q}\left[\log\frac{\pi_\phi(\boldsymbol{a}_{0:K}^{\mathrm{EE}}|\mathbf{H}_\mathrm{B})}{q(\boldsymbol{a}_{1:K}^{\mathrm{EE}}|\boldsymbol{a}_0^{\mathrm{EE}},\mathbf{H}_\mathrm{B})}\right].
\end{equation}

\subsubsection{Spatial-Temporal Robot Graph}
The physical architecture impacts the motion of the whole robot, determining whether it can complete the task.
Meanwhile, the historical spatial information is also important to future movements.
Thus, we propose a spatial-temporal graph method to model the robot architecture at each step and the robot motion at continuous timesteps, representing the static spatial information and dynamic motion features.

\vspace{-4mm}
\paragraph{Spatial Structure Graph Construction.}
To represent the robot structure, we first parse the Unified Robot Description Format (URDF) file which is usually used to describe the static physical structure of robot such as joint types, joint limits, and link lengths.
Then, we define a dual-arm system as an undirected graph $G_\mathrm{S}=\langle V_\mathrm{S},E_\mathrm{S}\rangle$ based on the joint and link configuration.
Here, $V_\mathrm{S}=\{\boldsymbol{v}_{\mathrm{j}_1}, \boldsymbol{v}_{\mathrm{j}_2}, \ldots, \boldsymbol{v}_{\mathrm{j}_m}\}$ and $E_\mathrm{S}=\{\boldsymbol{e}_{ij}|(\text{joint}_i,\text{joint}_j)\in \text{links}\}$ represents the node set of joints and edge set of links respectively, where $m$ denotes the number of joint.
We use $\boldsymbol{f}_i\in\mathbb{R}^D$ to denote the value of $i$-$th$ node feature.
It consists of three attributes as follows:
\begin{itemize}
    \item \textit{Joint Coordinate}: we use a vector in Cartesian coordinate system, $\boldsymbol{f}_i^\mathrm{JC}=[\boldsymbol{x}_i,\boldsymbol{y}_i,\boldsymbol{z}_i]\in\mathbb{R}^3$, to denote the absolute coordinates of the $i$-$th$ joint. The vector is normalized according to the workspace boundary for stable convergence of model training.
    \item \textit{Joint Distance}: To measure the spatial relationship between the node $\boldsymbol{v}_i$ and the other nodes $\boldsymbol{v}_j$, we compute the Euclidean Distance between $\boldsymbol{v}_i$ and $\boldsymbol{v}_j$:
    \vspace{-1mm}
    \begin{equation}
        \boldsymbol{f}_i^\mathrm{JD}=||\boldsymbol{v}_i-\boldsymbol{v}_j||_2,
        \vspace{-1mm}
    \end{equation}
    where $\boldsymbol{f}_i^\mathrm{JD}\in\mathbb{R}^{m}$ and $||\cdot||_2$ means the Euclidean norm.
    \item \textit{Body Label}: To discriminate the source of node $\boldsymbol{v}_i$, we use a one-hot vector $\boldsymbol{f}^\mathrm{BL}_i=[0,1]$ as one of its features. It can also help the policy capture the motion modes of different robot arms such as the symmetry.
\end{itemize}
\vspace{1mm}
We concatenate $\boldsymbol{f}_i^\mathrm{JC}$, $\boldsymbol{f}_i^\mathrm{JD}$, and  $\boldsymbol{f}_i^\mathrm{BL}$ to form the completed feature $\boldsymbol{f}_i\in\mathbb{R}^{D_f}$, where ${D_f}$ demotes the dimension of node features.

\vspace{-3mm}
\paragraph{Spatial-Temporal Graph Learning.}
Given the same instruction and observation, different historical robot poses result in different predictions.
Thus, we build the spatial structure graph with the temporal motion information.
Specifically, we build a spatial-temporal graph $G_\mathrm{ST}=\langle V_\mathrm{ST},E_\mathrm{ST}\rangle$ by combining $\{G_\mathrm{S}^i=\langle V^i_\mathrm{S}, E^i_\mathrm{S}\rangle\}_{i=0}^{T-1}$ from historical timesteps, where $T$ denotes the number of history steps.
In $G_\mathrm{ST}$, the node set $V_\mathrm{ST}$ contains $V^i_\mathrm{S}$ from historical static spatial graph.
Additionally, we add edges $E'$ linking the same joint node $\mathrm{joint}^t_i$ at different timesteps, in order to establish the correlation of joint motions at continuous times.
It can be formulated as follows:
\begin{equation}
    V_\mathrm{ST}=\cup \{V^i_\mathrm{S}\}_{i=0}^{T-1},\ \  E_\mathrm{ST}=\cup \{E^i_\mathrm{S}\}_{i=0}^{T-1} \cup E',
\end{equation}
\begin{equation}
    E'=\{\boldsymbol{e}_i^{t,t'}| t, t'\in\{0, 1, \ldots, T\}, t\neq t'\}.
\end{equation}

In this way, we obtain the whole spatial-temporal graph.
A Graph Convolutional Network (GCN) is then adopted to propagate and aggregate node features across the graph.
The GCN layer updates each node feature $\boldsymbol{f}_i$ by aggregating the features of its neighboring nodes, thereby capturing the relational and structural information of the robotic arms.
We use the node feature $\mathbf{H}_\mathrm{G}$ of the last encoder layer as the representation of the robot structure to condition the denoising process.

\subsubsection{Kinematics Regularizer}
To control the end-effector effectively, the generated pose trajectory must be processed by an Inverse Kinematics (IK) solver, which calculates the joint configurations to achieve the specified poses. 
However, because the predicted trajectory is generated without considering the robot kinematic constraints, it often falls outside the IK solver’s feasible range, resulting in high failure rates during execution. 
To address this limitation, we propose a kinematics regularizer into the end-effector pose learning objective. 
This regularizer aligns the predicted poses with the robot kinematic constraints, ensuring that the generated trajectory remains within the solvable space of the IK solver, thus enhancing then reliability of trajectory execution.

\vspace{-4mm}
\paragraph{Differentiable Kinematics.}
Given a joint configuration $\Theta = [\theta_1, \theta_2, \ldots, \theta_n]$, the corresponding end-effector pose $\mathbf{T} \in SE(3)$ can be computed using forward kinematics, represented as a mapping $f(\Theta): \mathbb{R}^n \rightarrow SE(3)$. 
This mapping from joint space to end-effector space is differentiable, namely Differentiable Forward Kinematics (DFK), enabling the use of gradients to optimize our control policy.
Leveraging DFK, our policy learns to predict the next joint configuration $\boldsymbol{\hat{a}}^\mathrm{joint}$, from which we compute an intermediate end-effector pose $\mathbf{H}_{\mathrm{R}}$. 
By using $\mathbf{H}_{\mathrm{R}}$ as a reference, we guide a denoising process to generate a precise and executable end-effector pose.

Specifically, we combine structure features $\mathbf{H}_\mathrm{G}$ with the last hidden state $\mathbf{H}_\mathrm{B}$, projecting to the joint space and using the DFK to obtain reference as $\mathbf{H}_\mathrm{R}$ follows:
\begin{equation}
    \boldsymbol{\hat{a}}^\mathrm{joint} = {\operatorname{Proj}([\mathbf{H}_\mathrm{B}, \mathbf{H}_\mathrm{G}])},\ \ \mathbf{H}_{\mathrm{R}} = \operatorname{DFK}(\boldsymbol{\hat{a}}^\mathrm{joint}).
\end{equation}
To ensure the consistency between predicted and actual joint angles, we minimize the joint loss:
\begin{equation}
    \mathcal{L}_\mathrm{joint} = \mathbb{E}_{\boldsymbol{a}^\mathrm{joint}\sim q(\boldsymbol{a}_0)}\left[||\boldsymbol{a}^\mathrm{joint} - \boldsymbol{\hat{a}}^\mathrm{joint}||^2\right].
\end{equation}

\vspace{-5mm}
\paragraph{Conditioning Diffusion Process on Kinematics.}
To enforce kinematic consistency, we condition the diffusion process on the reference representation $\mathbf{H}_\mathrm{R}$, an auxiliary input encoding kinematic constraints. 
This allows the predicted pose trajectory to stay within feasible space.
Given the diffusion steps from Eq.(\ref{eq:original_diffusion_action}) and Eq.(\ref{eq:original_diffusion_action_prob}), we have:
% \vspace{-1mm}
\begin{equation}
    \boldsymbol{a}_{k-1}^{\mathrm{EE}} \sim \pi_\phi(\boldsymbol{a}_{k}^{\mathrm{EE}}|\boldsymbol{a}_{k-1}^{\mathrm{EE}}, \mathbf{H}_\mathrm{B}, \mathbf{H}_\mathrm{R}),
\end{equation}
\begin{equation}
    \begin{split}
        &\pi_\phi(\boldsymbol{a}_{k-1}^{\mathrm{EE}}|\boldsymbol{a}_k^{\mathrm{EE}},\mathbf{H}_\mathrm{B}, \mathbf{H}_\mathrm{R})=\\
        &\mathcal{N}(\boldsymbol{a}_{k-1}^{\mathrm{EE}};\sqrt{\overline{\alpha}_{k-1}}\ \pi_\phi(\boldsymbol{a}_k^{\mathrm{EE}}, \mathbf{H}_\mathrm{B}, \mathbf{H}_\mathrm{R});(1-\overline{\alpha}_{k-1})\boldsymbol{I}).
    \end{split}
\end{equation}
Incorporating DFK into the diffusion process allows gradients from the pose loss to propagate back through the kinematic function, ensuring that each denoising step maintains compliance with joint constraints, thereby optimizing the end-effector’s control accuracy and robustness.

\subsection{Training and Inference}
\paragraph{Training.}
We use the conditional action generation mode to train KStar Diffuser, which is formulated as a conditional denoising diffusion.
The loss function is defined as the Mean Square Error (MSE) as follows:
\vspace{-1mm}
\begin{equation}
    \boldsymbol{a}_k = \sqrt{\overline{\alpha}_k}\,\boldsymbol{a}_0+\sqrt{1-\overline{\alpha}_k}\,\epsilon,
\end{equation}
\begin{equation}
    \mathcal{L}_\mathrm{EE} = \mathbb{E}_{\boldsymbol{a}_0^{\mathrm{EE}}\sim q(\boldsymbol{a}_0),k}\left[||\boldsymbol{a}_0^{\mathrm{EE}}-\pi_{\phi}(\boldsymbol{a}_k^{\mathrm{EE}}, k, C)||^2\right],
\end{equation}
\begin{equation}
    \mathcal{L}_\mathrm{joint} = \mathbb{E}_{\boldsymbol{a}^\mathrm{joint}_0\sim q(\boldsymbol{a}_0)}[||\boldsymbol{a}_0^\mathrm{joint} - \pi_\phi(\mathbf{H}_\mathrm{B}, \mathbf{H}_\mathrm{G})||^2],
\end{equation}
\begin{equation}
    \mathcal{L} = \lambda\,\mathcal{L}_\mathrm{EE} + (1-\lambda) \mathcal{L}_\mathrm{joint},
\end{equation}
where $\boldsymbol{a}_k$ is obtained by the forward diffusion process and $C$ is the combination of $\mathbf{H}_\mathrm{B}$, $\mathbf{H}_\mathrm{G}$, $\mathbf{H}_\mathrm{R}$. 
$\lambda$ is the trade-off coefficient.

\vspace{-4mm}
\paragraph{Inference.} 
Sampling from a Gaussian noise $\epsilon_k$, the policy $\pi_{\phi}$ performs $K$ iterations to gradually denoise a random noise $\epsilon_k$ into the noise-free action $\boldsymbol{a}_0$:
\begin{equation}
    \boldsymbol{a}^\mathrm{EE}_{k-1}=\sqrt{\overline{\alpha}_{k-1}}\,\pi_{\phi}(\boldsymbol{a}^\mathrm{EE}_k, k, C)+\sqrt{1-\overline{\alpha}_{k-1}}\,\epsilon_k.
\end{equation}
We use the predicted $\boldsymbol{a}^\mathrm{EE}_0$ as the final action to control robot execution.

\section{Experiment}
\subsection{Dataset and Evaluation Settings}
\paragraph{Dataset.}
Bimanual manipulation tasks demand high levels of coordination, synchronization, and symmetry awareness between the two robotic arms, making them inherently more challenging than single-arm tasks. 
To assess the capabilities of KStar Diffuser in these areas, we conducted comprehensive experiments using the RLBench2 benchmark~\cite{grotz2024peract2}, an extended version of RLBench tailored for bimanual manipulation and comprising tasks closely resembling real-world scenarios. 
Please refer to Appendix~\ref{appendix:sec_task_details} for more details about RLBench2 and real-world tasks.

\begin{figure}[h]
    \vspace{-3mm}
    % \centering
    \includegraphics[width=\linewidth]{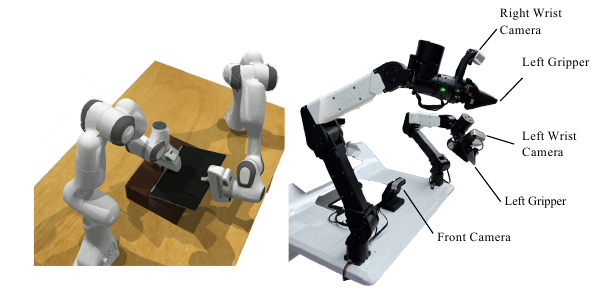}
    \vspace{-1mm}
    \caption{\textit{The left:} the simulation environment of \texttt{pick\_laptop} task. \textit{The right}: the ALOHA device used in the real-world tasks.}
    \label{fig:device}
    \vspace{-4mm}
\end{figure}

\vspace{-4mm}
\paragraph{Evaluation Settings.}
To evaluate the policy performance, we employ success rate as the primary metric. 
Although the policy generates multiple sequential actions during execution, we primarily focus on the final goal achievement rather than intermediate steps. 
Each task is associated with a specific success criterion defined by its target state. 
To comprehensively assess the policy's capability, we conduct experiments with varying numbers of demonstrations (20 and 100) during training. 
Figure~\ref{fig:device} illustrates our experimental setup, including both the simulation environment and the Cobot Agilex ALOHA robot. 
Detailed descriptions of simulation tasks and real-world experimental settings are provided in the Appendix~\ref{appendix:sec_implementation_details}.

\subsection{Baselines}
We systematically evaluate KStar Diffuser against state-of-the-art methods in two major categories:
\vspace{-4mm}
\paragraph{Transformer-based methods:}
(1) Action Chunking with Transformers (\textbf{ACT})~\cite{zhao2023learning} employs a Conditional VAE architecture, consisting of an encoder-decoder framework for joint angle sequence prediction;
(2) Robotic View Transformer Leader Following (\textbf{RVT-LF})~\cite{grotz2024peract2} leverages RVT~\cite{goyal2023rvt} as its backbone, incorporating a multi-view transformer for cross-view information aggregation and image re-rendering, coupled with a leader following mechanism for action prediction;
(3) Perceiver-Actor Leader Following (\textbf{PerAct-LF})~\cite{grotz2024peract2} adopts the leader following paradigm based on PerAct~\cite{shridhar2023perceiver}, utilizing a perceiver Transformer to encode both instructions and voxel observations for optimal voxel action generation;
(4) \textbf{PerAct2}~\cite{grotz2024peract2} enhances PerAct by implementing a unified feature space for dual-arm actions and employing combined self-attention for synchronized bimanual action prediction.

\vspace{-4mm}
\paragraph{Diffusion-based methods:}
(1) Joint-based Diffusion Policy (\textbf{DP-J})~\cite{chi2023diffusion} adopts a diffusion model to robotic manipulation within an imitation learning framework, focusing on joint angle prediction;
(2) EndEffector-based Diffusion Policy (\textbf{DP-EE}) is a DP variant which we reimplement Diffusion Policy to predict end-effector poses instead of joint angles, offering an alternative control paradigm;
(3) 3D Diffusion Policy (\textbf{DP3})~\cite{ze20243d} enhances 3D perception by incorporating point clouds for joint angle prediction.

\begin{table*}[ht]
  \centering
  \caption{The experimental result on simulated tasks. We train the policy with the setting of different training demonstrations, \ie $[20, 100]$, to test its capability comprehensively with 100 times each task. The best results are in \textbf{bold}. Each result is reported with three seeds on average.}
  \resizebox{0.90\linewidth}{!}{
    \begin{tabular}{lc>{\raggedleft\arraybackslash}p{1.6cm}>{\raggedleft\arraybackslash}p{1.6cm}>{\raggedleft\arraybackslash}p{1.7cm}>{\raggedleft\arraybackslash}p{1.6cm}>{\raggedleft\arraybackslash}p{1.6cm}>{\raggedleft\arraybackslash}p{1.6cm}}
    \toprule[1pt]
          Method & Backbone & \multicolumn{1}{r}{\makecell{Push\\Box}} & \multicolumn{1}{c}{\makecell{Lift\\Ball}} & \multicolumn{1}{r}{\makecell{Handover\\Item (easy)}} & \multicolumn{1}{r}{\makecell{Pick\\Laptop}} & \multicolumn{1}{r}{\makecell{Sweep\\Dustpan}} & \multicolumn{1}{c}{Overall} \\ [0.5ex]
    \midrule
    \midrule
    \multicolumn{1}{l}{\textit{\textbf{20 demos}}} &       &       &       &       &       &       &  \\
    \midrule
    ACT~\cite{zhao2023learning}  & Transformer & 26.0\textcolor[gray]{0.5}{\small{$\ \,\pm4.6$}} & 17.3\textcolor[gray]{0.5}{\small{$\ \,\pm5.8$}} & 0.0\textcolor[gray]{0.5}{\small{$\ \,\pm0.0$}} & 0.0\textcolor[gray]{0.5}{\small{$\ \,\pm0.0$}} & 0.0\textcolor[gray]{0.5}{\small{$\ \,\pm0.0$}} & 8.7\textcolor[gray]{0.5}{\small{$\ \,\pm1.6$}} \\
    RVT-LF~\cite{grotz2024peract2} & Transformer & 44.0\textcolor[gray]{0.5}{\small{$\ \,\pm1.0$}} & 12.3\textcolor[gray]{0.5}{\small{$\ \,\pm6.7$}} & 0.0\textcolor[gray]{0.5}{\small{$\ \,\pm0.0$}} & 2.0\textcolor[gray]{0.5}{\small{$\ \,\pm1.7$}} & 0.0\textcolor[gray]{0.5}{\small{$\ \,\pm0.0$}} & 11.7\textcolor[gray]{0.5}{\small{$\ \,\pm1.5$}} \\
    PerAct-LF~\cite{grotz2024peract2} & Transformer & 63.7\textcolor[gray]{0.5}{\small{$\ \,\pm3.1$}} & 42.3\textcolor[gray]{0.5}{\small{$\ \,\pm8.1$}} & 3.3\textcolor[gray]{0.5}{\small{$\ \,\pm0.6$}} & 7.0\textcolor[gray]{0.5}{\small{$\ \,\pm2.7$}} & 4.7\textcolor[gray]{0.5}{\small{$\ \,\pm0.6$}} & 24.2\textcolor[gray]{0.5}{\small{$\ \,\pm1.6$}} \\
    PerAct2~\cite{grotz2024peract2} & Transformer & 59.7\textcolor[gray]{0.5}{\small{$\ \,\pm7.6$}} & 37.7\textcolor[gray]{0.5}{\small{$\ \,\pm4.7$}} & 8.0\textcolor[gray]{0.5}{\small{$\ \,\pm1.0$}} & 17.3\textcolor[gray]{0.5}{\small{$\ \,\pm0.6$}} & 1.7\textcolor[gray]{0.5}{\small{$\ \,\pm1.5$}} & 24.9\textcolor[gray]{0.5}{\small{$\ \,\pm2.1$}} \\
    \midrule
    DP-J~\cite{chi2023diffusion}  & Diffusion & 30.0\textcolor[gray]{0.5}{\small{$\ \,\pm3.6$}} & 19.7\textcolor[gray]{0.5}{\small{$\ \,\pm2.1$}} & 0.0\textcolor[gray]{0.5}{\small{$\ \,\pm0.0$}} & 0.0\textcolor[gray]{0.5}{\small{$\ \,\pm0.0$}} & 0.0\textcolor[gray]{0.5}{\small{$\ \,\pm4.0$}} & 10.0\textcolor[gray]{0.5}{\small{$\ \,\pm1.0$}} \\
    DP-EE~\cite{chi2023diffusion} & Diffusion & 34.7\textcolor[gray]{0.5}{\small{$\ \,\pm2.5$}} & 34.3\textcolor[gray]{0.5}{\small{$\ \,\pm7.4$}} & 0.7\textcolor[gray]{0.5}{\small{$\ \,\pm0.6$}} & 1.3\textcolor[gray]{0.5}{\small{$\ \,\pm0.6$}} & 14.0\textcolor[gray]{0.5}{\small{$\ \,\pm1.7$}} & 17.0\textcolor[gray]{0.5}{\small{$\ \,\pm1.6$}} \\
    DP3~\cite{ze20243d} & Diffusion & 25.3\textcolor[gray]{0.5}{\small{$\ \,\pm4.0$}} & 34.3\textcolor[gray]{0.5}{\small{$\ \,\pm7.6$}} & 0.0\textcolor[gray]{0.5}{\small{$\ \,\pm0.0$}} & 2.7\textcolor[gray]{0.5}{\small{$\ \,\pm2.9$}} & 0.0\textcolor[gray]{0.5}{\small{$\ \,\pm0.0$}} & 12.5\textcolor[gray]{0.5}{\small{$\ \,\pm1.3$}} \\
    \rowcolor{blue!5!cyan!10}KStar Diffuser (Ours) & Diffusion & 79.3\textcolor[gray]{0.5}{\small{$\ \,\pm3.5$}} & 87.0\textcolor[gray]{0.5}{\small{$\ \,\pm2.7$}} & 23.7\textcolor[gray]{0.5}{\small{$\ \,\pm0.6$}} & 17.0\textcolor[gray]{0.5}{\small{$\ \,\pm2.0$}} & 83.0\textcolor[gray]{0.5}{\small{$\ \,\pm4.4$}} & 58.0\textcolor[gray]{0.5}{\small{$\ \,\pm1.4$}} \\
    \midrule[0.75pt]
    \multicolumn{8}{l}{\textit{\textbf{100 demos}}} \\
    \midrule
    \midrule
    ACT~\cite{zhao2023learning}   & Transformer & 48.7\textcolor[gray]{0.5}{\small{$\ \,\pm6.8$}} & 40.7\textcolor[gray]{0.5}{\small{$\ \,\pm2.1$}} & 0.0\textcolor[gray]{0.5}{\small{$\ \,\pm0.0$}} & 0.0\textcolor[gray]{0.5}{\small{$\ \,\pm0.0$}} & 0.0\textcolor[gray]{0.5}{\small{$\ \,\pm0.0$}} & 17.9\textcolor[gray]{0.5}{\small{$\ \,\pm1.0$}} \\
    RVT-LF~\cite{grotz2024peract2} & Transformer & 76.3\textcolor[gray]{0.5}{\small{$\ \,\pm1.5$}} & 28.7\textcolor[gray]{0.5}{\small{$\ \,\pm1.0$}} & 0.0\textcolor[gray]{0.5}{\small{$\ \,\pm0.0$}} & 1.3\textcolor[gray]{0.5}{\small{$\ \,\pm1.2$}} & 2.3\textcolor[gray]{0.5}{\small{$\ \,\pm3.2$}} & 21.7\textcolor[gray]{0.5}{\small{$\ \,\pm0.7$}} \\
    PerAct-LF~\cite{grotz2024peract2} & Transformer & 66.3\textcolor[gray]{0.5}{\small{$\ \,\pm3.1$}} & 68.3\textcolor[gray]{0.5}{\small{$\ \,\pm6.1$}} & 7.3\textcolor[gray]{0.5}{\small{$\ \,\pm2.1$}} & 14.3\textcolor[gray]{0.5}{\small{$\ \,\pm3.8$}} & 8.0\textcolor[gray]{0.5}{\small{$\ \,\pm2.7$}} & 32.9\textcolor[gray]{0.5}{\small{$\ \,\pm2.5$}} \\
    PerAct2~\cite{grotz2024peract2} & Transformer & 77.0\textcolor[gray]{0.5}{\small{$\ \,\pm3.0$}} & 47.7\textcolor[gray]{0.5}{\small{$\ \,\pm3.2$}} & 10.3\textcolor[gray]{0.5}{\small{$\ \,\pm0.6$}} & 34.3\textcolor[gray]{0.5}{\small{$\ \,\pm4.0$}} & 6.7\textcolor[gray]{0.5}{\small{$\ \,\pm2.1$}} & 33.9\textcolor[gray]{0.5}{\small{$\ \,\pm1.0$}} \\
    \midrule
    DP-J~\cite{chi2023diffusion}  & Diffusion & 62.7\textcolor[gray]{0.5}{\small{$\ \,\pm7.6$}} & 43.3\textcolor[gray]{0.5}{\small{$\ \,\pm2.1$}} & 0.0\textcolor[gray]{0.5}{\small{$\ \,\pm0.0$}} & 1.3\textcolor[gray]{0.5}{\small{$\ \,\pm0.6$}} & 0.0\textcolor[gray]{0.5}{\small{$\ \,\pm4.0$}} & 21.5\textcolor[gray]{0.5}{\small{$\ \,\pm1.6$}} \\
    DP-EE~\cite{chi2023diffusion} & Diffusion & 61.0\textcolor[gray]{0.5}{\small{$\ \,\pm2.0$}} & 59.7\textcolor[gray]{0.5}{\small{$\ \,\pm3.5$}} & 2.0\textcolor[gray]{0.5}{\small{$\ \,\pm1.0$}} & 10.3\textcolor[gray]{0.5}{\small{$\ \,\pm3.1$}} & 69.7\textcolor[gray]{0.5}{\small{$\ \,\pm2.5$}} & 40.5\textcolor[gray]{0.5}{\small{$\ \,\pm1.2$}} \\
    DP3~\cite{ze20243d} & Diffusion & 56.0\textcolor[gray]{0.5}{\small{$\ \,\pm3.6$}} & 64.0\textcolor[gray]{0.5}{\small{$\ \,\pm2.7$}} & 0.0\textcolor[gray]{0.5}{\small{$\ \,\pm0.0$}} & 6.3\textcolor[gray]{0.5}{\small{$\ \,\pm3.1$}} & 1.7\textcolor[gray]{0.5}{\small{$\ \,\pm2.1$}} & 25.6\textcolor[gray]{0.5}{\small{$\ \,\pm1.4$}} \\
    \rowcolor[rgb]{ .902,  .996,  1.0}KStar Diffuser (Ours) & Diffusion & \textbf{83.0}\textcolor[gray]{0.5}{\small{$\ \,\pm1.7$}} & \textbf{98.7}\textcolor[gray]{0.5}{\small{$\ \,\pm1.5$}} & \textbf{27.0}\textcolor[gray]{0.5}{\small{$\ \,\pm1.7$}} & \textbf{43.7}\textcolor[gray]{0.5}{\small{$\ \,\pm4.5$}} & \textbf{89.0}\textcolor[gray]{0.5}{\small{$\ \,\pm5.2$}} & \textbf{68.2}\textcolor[gray]{0.5}{\small{$\ \,\pm2.1$}} \\
    \bottomrule[1pt]
    \end{tabular}
    }%
  \label{tab:simulation_performance}%
  \vspace{-1mm}
\end{table*}%

\subsection{Comparison Results with SOTA Methods}
\paragraph{Experimental Results on RLBench2.}
As shown in Table~\ref{tab:simulation_performance}, the KStar Diffuser significantly outperforms other state-of-the-art baselines, achieving more than $20\%$ higher overall performance with both 20 and 100 training demonstrations.
We found that:

\textbf{(1) Similar to learning a single-arm policy}, the process of learning a bimanual policy can adapt quickly and achieve a high success rate, given a relatively consistent distribution of task trajectories.
In \texttt{push\_box}, where the objective is for both arms to push a box toward a specified target along fixed trajectories, our KStar Diffuser and other baseline models perform well.
However, as task complexity increases, success rates decrease. 
For example, in the \texttt{lift\_ball} task, both arms must lift a large ball \textit{simultaneously} to complete the task. 
Any asynchrony in movement can cause instability, leading to the ball slipping and ultimately resulting in task failure.
Our KStar Diffuser achieves its robust performance on such bimanual tasks by \textbf{explicitly modeling the spatial and motion relationships between the two arms}, surpassing other methods more than 6\%.

\begin{figure*}[h]
    \includegraphics[width=\linewidth]{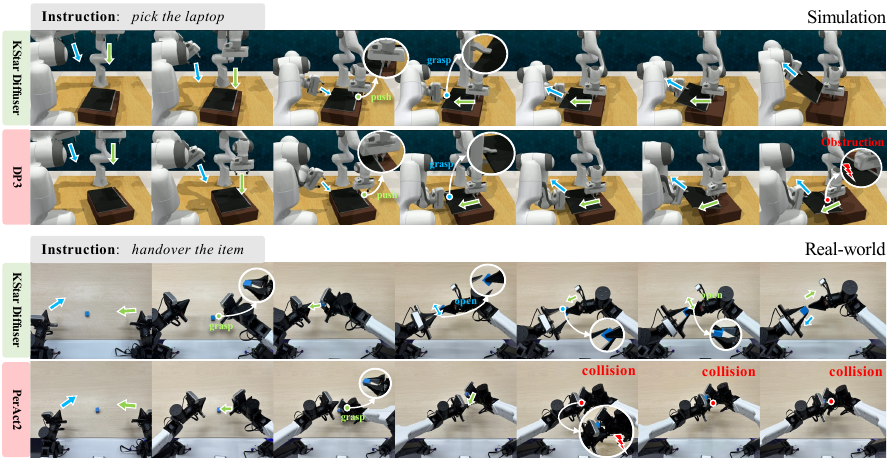}
    \caption{The visualization of bimanual manipulation on simulated RLBench2 and real-world tasks. The {\color[RGB]{0,176,240} blue} annotations represent the motion of the robot's left arm, while the {\color[RGB]{146,220,80} green} annotations indicate the motion of the right arm.}
    \label{fig:visualization}
\end{figure*}

\textbf{(2) Distinct from single-arm systems}, bimanual robotic systems possess the capability for \textbf{collaborative manipulation}.
Methods which are directly adapted from single-arm to bimanual manipulation exhibit high failure rates in tasks, \eg,  \texttt{pick\_laptop}, as they lack consideration of the spatial and motion relationships between the arms.
Specifically, as shown in Figure~\ref{fig:visualization}, this task involves picking up a notebook lying flat on a cabinet surface. 
Given that the notebook rests fully against the tabletop, direct grasping by the robotic arm is not possible. 
Instead, an effective strategy is to control one arm to push the notebook outward from the cabinet by a short distance, allowing the other arm to pick it up. 
KStar Diffuser achieves a success rate approximately 9\% higher than other methods, demonstrating its ability to \textbf{capture the coordinated motion patterns} required for collaborative object manipulation between two arms.

\vspace{-4mm}
\paragraph{Real-world Experimental Results.}
To comprehensively evaluate the policy's effectiveness, we build 2 tasks in real-world based on the simulation benchmark.
The performances on real-world tasks are illustrated in Table~\ref{tab:real_performance}.

Similar to the simulation results, we observe that policies which do not consider about bimanual scenarios, \ie, ACT, DP, and DP3, demonstrate limited capability across all bimanual tasks, achieving around 20\% success ratio on average.
Although PerAct2 is designed for bimanual tasks via mapping bimanual actions into a shared learning space, it fails to capture the spatial structure of the bimanual system, leading to ineffective arm coordination during execution.
Furthermore, we also found PerAct2 faces \textit{significant inverse kinematics issues} with its predicted end-effector poses, including joint configuration conflicts and unreachable positions which are shown in Figure~\ref{fig:visualization}.
It is likely due to PerAct2's limited ability to capture the complex spatial constraints and kinematic relationships within the bimanual robotic system.
In contrast, KStar Diffuser achieves superior bimanual coordination, surpassing other methods by over 10\%, as it successfully captures the motion patterns between dual arms and predicts feasible end-effector poses.

\begin{table}[ht]
  \centering
  \caption{The result of real-world tasks. We train all policies with 100 demostrations and test 15 times. The best result are in \textbf{bold}.}
  \resizebox{0.94\linewidth}{!}{
    \begin{tabular}{l|rrr}
    \toprule[1pt]
    Methods      & \multicolumn{1}{c}{Lift Plate} & \multicolumn{1}{c}{Handover} & \multicolumn{1}{c}{Overall} \\
    \midrule\midrule
    ACT~\cite{zhao2023learning}   & 37.8\textcolor[gray]{0.5}{\small{$\ \,\pm8.3$}}       & 0.0\textcolor[gray]{0.5}{\small{$\ \,\pm0.0$}}       & 18.9\textcolor[gray]{0.5}{\small{$\ \,\pm11.2$}}   \\
    DP~\cite{chi2023diffusion}    & 42.1\textcolor[gray]{0.5}{\small{$\ \,\pm6.5$}}       & 0.0\textcolor[gray]{0.5}{\small{$\ \,\pm0.0$}}       & 21.1\textcolor[gray]{0.5}{\small{$\ \,\pm12.5$}}   \\
    DP3~\cite{ze20243d}  & 44.0\textcolor[gray]{0.5}{\small{$\ \,\pm8.3$}}       & 0.0\textcolor[gray]{0.5}{\small{$\ \,\pm0.0$}}       & 22.0\textcolor[gray]{0.5}{\small{$\ \,\pm13.4$}}      \\
    PerAct2~\cite{grotz2024peract2} & 51.1\textcolor[gray]{0.5}{\small{$\ \,\pm6.3$}}       & 8.8\textcolor[gray]{0.5}{\small{$\ \,\pm3.0$}}       & 29.9\textcolor[gray]{0.5}{\small{$\ \,\pm10.2$}}    \\
    \rowcolor{blue!5!cyan!10}KStar Diffuser (Ours)  & \textbf{66.7}\textcolor[gray]{0.5}{\small{$\ \,\pm5.3$}}       & \textbf{19.7}\textcolor[gray]{0.5}{\small{$\ \,\pm5.3$}}       & \textbf{43.1}\textcolor[gray]{0.5}{\small{$\ \,\pm17.8$}}        \\
    \bottomrule[1pt]
    \end{tabular}%
  \label{tab:real_performance}}%
  \vspace{-.5mm}
\end{table}%

\begin{table}[htbp]
  \centering
  \caption{Ablation study of model components.}
  \resizebox{0.92\linewidth}{!}{
    \begin{tabular}{cc|rr|r}
    \toprule[1pt]
          ST Graph & KR & \multicolumn{1}{r}{Handover-S} & \multicolumn{1}{r}{Handover-R} & \multicolumn{1}{|c}{Overall} \\
    \midrule\midrule
    \rowcolor{blue!5!cyan!10}$\checkmark$ & $\checkmark$ & \textbf{27.0}\textcolor[gray]{0.5}{\small{$\ \,\pm1.7$}}       & \textbf{19.7}\textcolor[gray]{0.5}{\small{$\ \,\pm5.3$}}      & \textbf{23.4}\textcolor[gray]{0.5}{\small{$\ \,\pm5.2$}}  \\
    $\checkmark$ & $\times$& 18.3\textcolor[gray]{0.5}{\small{$\ \,\pm1.5$}}       & 15.3\textcolor[gray]{0.5}{\small{$\ \,\pm3.3$}}      & 16.8\textcolor[gray]{0.5}{\small{$\ \,\pm2.1$}}  \\
     $\times$& $\times$ & 16.3\textcolor[gray]{0.5}{\small{$\ \,\pm3.1$}}       & 13.3\textcolor[gray]{0.5}{\small{$\ \,\pm9.4$}}      & 14.8\textcolor[gray]{0.5}{\small{$\ \,\pm2.1$}} \\
    \bottomrule[1pt]
    \end{tabular}%
  \label{tab:ablation}}%
\end{table}%

\subsection{Ablation Studies}
\paragraph{Effects of Model Components.}
To systematically evaluate the contribution of each component in KStar Diffuser, we conduct ablation experiments on the \texttt{handover\_item} task in both simulated and real-world environments. We design a progressive ablation process by first removing the Differential Forward Kinematics module while retaining the Spatial-Temporal Graph (ST Graph), and then completely disabling both ST Graph and Kinematics Regularizer (KR).

The experimental results in Table~\ref{tab:ablation} demonstrate the crucial role of each component.
The removal of KR leads to a significant decrease in success rate, particularly pronounced in real-world scenarios. 
This performance degradation can be attributed to the fundamental differences between simulated and real environments. 
While simulated environments maintain consistent and noise-free inputs, real-world scenarios introduce various perturbations, \eg, sensor noise and light reflection, making the policy more susceptible to kinematic constraint violations without the regularization effect of KR.
Further ablation by removing both ST Graph and KR results in a substantial performance drop across all experimental settings. 
This observation illustrates two key aspects: First, the ST Graph effectively captures the spatial-temporal dependencies among joints, which is essential for coordinating the relative positioning and interactions between robotic arms. 
Second, the graph structure's explicit encoding of the robotic physical architecture enhances the policy's robustness against unexpected perturbations by maintaining spatial and temporal coherence.
We conduct extensive ablation studies of action chunking size, history lengths and trade-off coefficient, please refer to Appendix~\ref{appendix:sec_extensive_ablation} for more details.

\vspace{-1mm}
\subsection{Qualitative Analysis}
We further present qualitative analysis in Figure~\ref{fig:visualization}.
We compare the performance of KStar Diffuser with DP3 and PerAct2 in executing bimanual manipulation tasks within both simulated and real-world environments, respectively.

In the simulation task, since the laptop rests flat on the cabinet, direct lifting is not feasible. 
One robotic arm initiates a forward push, creating space, while the other arm concurrently grasps and elevates the laptop. 
KStar Diffuser effectively models this dual-arm coordination, generating a precise trajectory of synchronized actions. 
Conversely, DP3, adapted from a single-arm policy to a dual-arm configuration, fails to achieve effective coordination.
Concretely, after executing the push motion, the right arm does not halt, obstructing the left arm's lifting process.

In the real-world task, KStar Diffuser generates an executable item transfer trajectory between the left and right arms, with no collisions throughout the task, reflecting its strong environmental adaptability and collision avoidance capabilities. 
Conversely, PerAct2 encounters collisions during the handover process (marked in red), indicating less effective handling of dynamic real-world variables and a lack of kinematic awareness on the robot movements.
More qualitative analysis can be found in Appendix~\ref{appendix:qualitative_analysis}.

\section{Conclusion}
In this paper, we have proposed a novel Kinematics enhanced Spatial-TemporAl gRaph Diffuser (KStar Diffuser) that explicitly incorporates both robot structures and kinematics into the bimanual motion generation process. 
It consists of a spatial-temporal robot graph that explicitly models the robot physical configuration to guide the generative action denoising procedure, and a kinematics regularizer that augments the NBP learning objective by introducing joint-space supervision.
Extensive experiments demonstrate that KStar Diffuser outperforms the baselines by a large margin on both simulation and real-world tasks.
\vspace{-4mm}
\paragraph{Limitations and Future Directions.}
While we explored robot structure impacts through GNN modeling and kinematic constraints, the core control logic of end-effector pose prediction and inverse kinematics remains. 
In the future, we aim to leverage neural networks to directly model joint movements, aligning the robot motion space with the Cartesian space of the human world.

{
    \small
    \bibliographystyle{ieeenat_fullname}
    \bibliography{main}
}

\input{X_suppl}

\end{document}

%% file: X_suppl.tex
\clearpage
\setcounter{section}{0}
\setcounter{figure}{0}
\setcounter{table}{0}
\setcounter{footnote}{0}
\renewcommand{\thesection}{\Alph{section}}
\renewcommand{\thetable}{\Alph{table}}
\renewcommand{\thefigure}{\Alph{figure}}

\maketitlesupplementary

We provide a more comprehensive \textbf{tasks descriptions} in Section~\ref{appendix:sec_task_details}, encompassing both simulated environments and real-world scenarios. 
In Section~\ref{appendix:sec_implementation_details}, we elaborate on the \textbf{implementation details}, including code base of baseline methods, and hyperparameter configurations for the backbone, graph encoder modules and optimization process. 
Notably, to thoroughly validate efficacy of KStar Diffuser, we conduct \textbf{extensive ablation studies} analyzing the impact of demonstration quantity, action chunking size, observation history length, and the control learning objective coefficient $\lambda$. 
The result is presented in Section~\ref{appendix:sec_extensive_ablation}.
Finally, we present more \textbf{qualitative analysis} in Section~\ref{appendix:qualitative_analysis}.

\section{Task Descriptions}\label{appendix:sec_task_details}
We conduct extensive experiments on both simulated tasks and real-world tasks to evaluate the effectiveness of our proposed KStar Diffuser.
Specifically, we selected five tasks from RLBench2, ranging from basic symmetrical tasks to advanced coordination-requiring tasks, including \texttt{push\_box}, \texttt{lift\_ball}, \texttt{handover\_item\_easy}, \texttt{sweep\_dustpan}, and \texttt{pick\_laptop}. 
For real-world evaluations, we created a similar setup with two bimanual tasks: \texttt{lift\_plate} and \texttt{handover\_item\_easy}.
We present the details about both simulated and real-world tasks in Table~\ref{tab:appendix_task_details}.
For each simulated task, we evaluate the model 100 times, whereas for each real-world task, we conduct 15 evaluations.

\vspace{-3mm}
\begin{table}[h]
  \centering
  \caption{Tasks Details.}
  \resizebox{\linewidth}{!}{
    \begin{tabular}{lccl}
    \toprule[1pt]
     Task & Duration & \# Keyframes & Instruction \\
     \midrule
     \multicolumn{4}{l}{\textit{\textbf{Simulated Tasks}}} \\
     \texttt{push\_box} & $4.33\mathrm{s}$ & $2.1$ & \textit{``Push the box to the red area.''} \\
     \texttt{lift\_ball} & $4.40\mathrm{s}$ & $4.0$ & \textit{``Lift the ball.''} \\
     \texttt{handover\_item\_easy} & $7.17\mathrm{s}$ & $7.5$ & \textit{``Handover the item.''} \\
     \texttt{sweep\_dustpan} & $4.93\mathrm{s}$ & $7.3$ & \textit{``Sweep the dust to the pan.''} \\
     \texttt{pick\_laptop} & $3.97\mathrm{s}$ & $7.2$  & \textit{``Pick up the notebook.''} \\
     \midrule
     \multicolumn{4}{l}{\textit{\textbf{Real-world Tasks}}} \\
     \texttt{lift\_plate} & $6.37\mathrm{s}$ & $3.4$ & \textit{``Lift the plate.''} \\
     \texttt{handover\_item\_easy} & $9.52\mathrm{s}$ & $8.6$ & \textit{``Handover the item.''} \\
    \bottomrule[1pt]
    \end{tabular}%
  \label{tab:appendix_task_details}
  }%
  \vspace{-1mm}
\end{table}%

\subsection{Simulated Tasks}\label{appendix:sec_simulated_task_details}
\begin{figure*}[h]
    \centering
    \includegraphics[width=\linewidth]{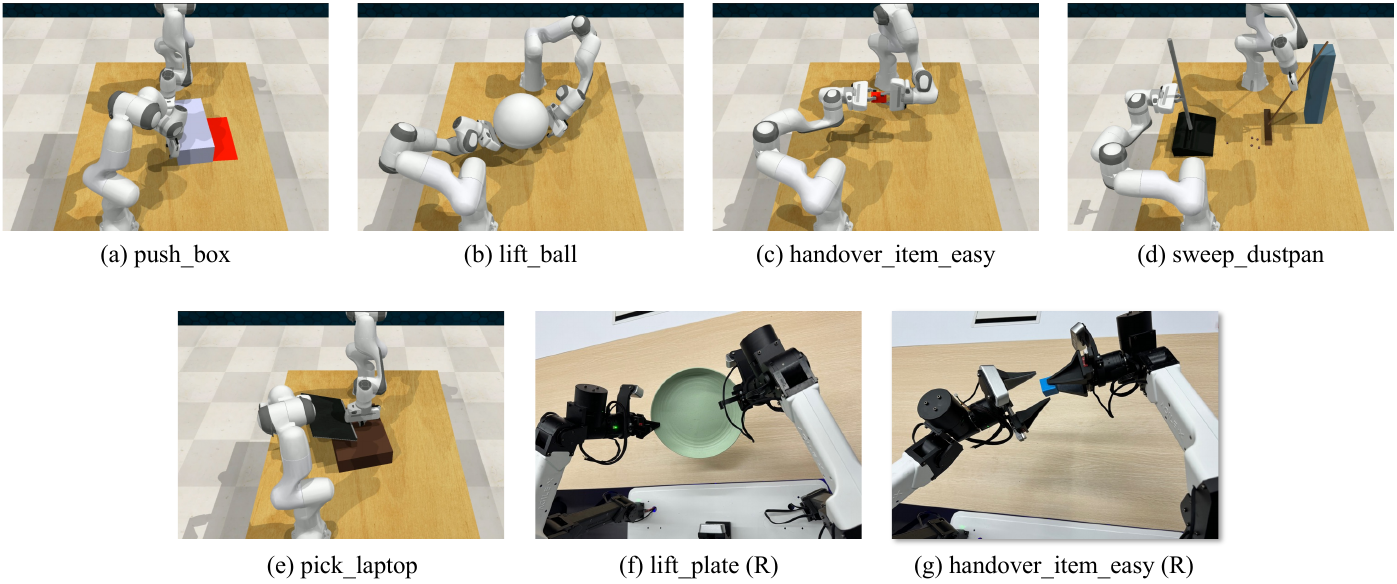}
    \caption{The visualization of simulated tasks and real-world tasks. The task with ``(R)'' means the real-world tasks.}
    \label{fig:appendix_tasks}
\end{figure*}

\paragraph{{\texttt{push\_box.}}}
As illustrated in Figure~\ref{fig:appendix_tasks}(a), the task requires the robot to utilize both arms to push a heavy box, weighing 50 kg, and transport it to a designated target area through a fix trajectory. 
The completion of the task is defined as successfully moving the box to the specified location. 
The scenario involves two key elements: a large box and a target area, with the primary challenge being the considerable weight of the box, which exceeds the capacity of a single arm to manage effectively. 
Notably, this task necessitates the use of both arms simultaneously, as a single robot is incapable of accomplishing it independently. 

\vspace{-4mm}
\paragraph{\textbf{\texttt{lift\_ball.}}}
As depicted in Figure~\ref{fig:appendix_tasks}(b), the task entails the robot using both arms to lift a large ball, achieving a minimum height of 0.95 meters to meet the success criteria. 
The object in this task is the large ball, presenting a significant coordination challenge. 
Due to the ball's size and the inability of the gripper to securely grasp it, the operation relies on coordinated non-prehensile manipulation, demanding precise synchronization of the arms during the lifting process. 
This task cannot be accomplished by a single robot because of the object's dimensions. 

\vspace{-4mm}
\paragraph{\textbf{\texttt{handover\_item\_easy.}}}
The task requires the robot to handover a red item by utilizing one arm to securely grasp and lift the item to a height of 80 cm while ensuring the other arm remains idle and unengaged, as shown in Figure~\ref{fig:appendix_tasks}(c). 
The object involved is a single red block, and the key challenge lies in coordinating the handover process effectively. 
Successful completion is determined when the item is accurately identified, grasped, and positioned at the required height with no actions performed by the idle arm. 

\vspace{-4mm}
\paragraph{\texttt{sweep\_dustpan.}}
It is shown in Figure~\ref{fig:appendix_tasks}(d) that the task involves the robot using a broom to sweep dust into a dust pan, requiring precise coordination of the sweeping motion to effectively collect the dust. 
The objects involved include a broom, a dust pan, supporting objects, and the dust itself. 
Successful completion is defined as all the dust being gathered inside the dust pan. 
The primary challenge lies in the accuracy and control of the sweeping motion to ensure that the dust is properly directed into the pan. 

\vspace{-4mm}
\paragraph{\texttt{pick\_laptop.}}
As illustrated in Figure~\ref{fig:appendix_tasks}(e), the task requires the robot to pick up a notebook placed on top of a block by first manipulating it into a position suitable for grasping. 
This involves performing non-prehensile actions, such as pushing or sliding, to adjust the notebook's orientation before securely grasping and lifting it off the block. 
The objects involved are a notebook and a block. Successful completion is defined as the robot lifting the notebook off the block.
Although the task can be performed with a single robotic arm, precise coordination is essential for effective manipulation. 

\subsection{Real-world Tasks}\label{appendix:sec_real-world_task_details}
\paragraph{\texttt{lift\_plate.}}
As shown in Figure~\ref{fig:appendix_tasks}(f), the task involves the robot using both arms to lift a plate, maintaining an elevated position for over 3 seconds to meet the success criteria. 
The target object is a plate requiring coordinated manipulation. 
Due to the plate's width and the need for stable control, the operation demands precise bimanual manipulation, requiring synchronized lifting motions from both arms. 
This task cannot be accomplished by a single robot arm given the plate's dimensions and stability requirements.

\vspace{-4mm}
\paragraph{\textbf{\texttt{handover\_item\_easy.}}}
Similar to the corresponding simulated task, this task requires the robot to handover a blue item by utilizing one arm to securely grasp and lift the item as shown in Figure~\ref{fig:appendix_tasks}(g).
However, the robot do not need to lift the item to a height of 80 cm, but 30cm instead considering the security, while ensuring the other arm remains idle and unengaged, 
The object involved is a blue block, and the key challenge is as the same as that of in simulation. 
Successful completion is determined when the item is accurately identified, grasped, and positioned at the required height with no actions performed by the idle arm. 

\section{Implementation Details}\label{appendix:sec_implementation_details}
\subsection{Details of KStar Diffuser}
\subsubsection{Backbone} 
\paragraph{Vision Branch.} Following the existing work~\cite{shridhar2023perceiver, grotz2024peract2}, we employ multiview RGB-D observation as visual input, where the resolution of RGB images is 3$\times$256$\times$256 and the depth data is processed into point clouds (65536$\times$3) using the camera's intrinsic and extrinsic parameters.
The camera views contain front, overhead, right\_over\_shoulder, left\_over\_shoulder, right\_wrist, left\_wrist.
We uniformly use a Vision Transformer which is trained from scratch as the vision encoder.

\vspace{-4mm}
\paragraph{Language Branch.} For the language instruction, we encode the instruction with CLIP's language encoder\footnote{https://huggingface.co/openai/clip-vit-base-patch32}. 
Specifically, the input sentence is preprocessed by the CLIP's tokenizer and then encoded to a sequence of dimensions $\mathbb{R}^{77\times512}$. 
We use its hidden state of ``\texttt{[CLS]}'' as the textual feature.
It is worth noting that we extracted all text features in advance. 
Therefore, throughout the training phase, the weights of the text encoder do not participate in the gradient calculation of backpropagation.

\vspace{-4mm}
\paragraph{Fusion Module.} The fusion module begins with upsampling the visual features, followed by feature-wise modulation (FiLM) to integrate textual features, projecting them into a high-dimensional semantic space. 
This hierarchical fusion process is performed iteratively $l_\mathrm{FiLM}$ times, where we empirically set $l_\mathrm{FiLM}$ to 3.

The relevant hyperparameters of the backbone are presented in Table~\ref{tab:appendix_backbone_hypers}.
\begin{table}[h]
  \centering
  \caption{The hyperparameters of Backbone.}
  % \resizebox{0.85\linewidth}{!}{
    \begin{tabular}{lrr}
    \toprule
     & \ \ \ Vision & \ \ \ Text \\
     \midrule
    patch size & $16$ & \texttt{N/A} \\
    hidden size   & $64$ & $512$ \\
    \# layers & $6$ & $12$ \\
    \# heads & $8$ &  $8$ \\
    intermediate size & $64$ & $2048$ \\
    dropout ratio & $0.1$ & $0.0$ \\
    activation & \texttt{lrelu} & \texttt{quick gelu} \\
    trainable & $\checkmark$ & $\times$ \\ 
    \bottomrule
    \end{tabular}%
  \label{tab:appendix_backbone_hypers}
  % }%
\end{table}%

\subsubsection{Spatial-Temporal Graph}
We construct the spatial graph based on the URDF file of robotic arms. In the simulation setup, we use two Franka Panda arms, each with 7 joints, resulting in a total of 14 joints. In the real-world setup, we employ a bimanual ALOHA device, which has 12 joints, with 6 joints per arm. Therefore, the number of nodes and edges in the spatial graph is set to 14 and 12 for the simulation, respectively, and 12 and 10 for the real-world setup.

For the dynamic spatio-temporal graph, we combine the spatial graphs of three consecutive timesteps and add inter-timestep edges connecting the same joint node across different timesteps. 
In simulated tasks, this results in a graph with 42 nodes and 33 edges. 
In real-world tasks, the graph has 36 nodes and 28 edges.

We use the Graph Convolutional Graph (GCN) encoder to obtain the spatial-temporal graph representation.
The detailed hyperparameters are presented in Table~\ref{tab:appendix_graph_hypers}.
\begin{table}[h]
  \centering
  \caption{The hyperparameters of GCN Encoder.}
  % \resizebox{0.85\linewidth}{!}{
    \begin{tabular}{lrr}
    \toprule
     & Simulation & Real-world \\
     \midrule
    \# nodes & $42$ & $36$ \\
    \# edges & $36$ & $28$ \\
    node dimension & $19$ & $19$ \\
    hidden size   & $128$ & $128$ \\
    intermediate size & $128$ & $128$ \\
    \# layers & $4$ & $4$  \\
    \bottomrule
    \end{tabular}%
  \label{tab:appendix_graph_hypers}
  % }%
\end{table}%

\subsubsection{Optimization Details}
During training, we follow the setup of Diffusion Policy, using the DDPM scheduler to forward and denoise, where the step of forward process and reverse process is set to 100 and 1, respectively.
Table~\ref{tab:appendix_train_hypers} shows the training hyperparamters. 
\vspace{-4mm}
\begin{table}[htbp]
    \centering
    \vspace{-4mm}
    \caption{The training hyperparameters.}
    \begin{tabular}{lc}
    \toprule
         & Values \\
             \midrule
        batch size & $64$ \\
        learning rate & $2e-4$ \\
        warmup step & $5$\,k \\
        weight decay & $1e-6$ \\
        lr scheduler & \texttt{cosine} \\
        training step & $150$\,k \\
        Optimizer & \texttt{AdamW} \\
    \bottomrule
    \end{tabular}
    \label{tab:appendix_train_hypers}
\end{table}

\subsection{Code Base}
The code bases employed for our evaluations are detailed as follows:
\begin{itemize}
    \item ACT: \href{https://github.com/markusgrotz/peract_bimanual}{https://github.com/markusgrotz/peract\_bimanual}
    \item RVT-LF: \href{https://github.com/markusgrotz/peract_bimanual}{https://github.com/markusgrotz/peract\_bimanual}
    \item PerAct-LF: \href{https://github.com/markusgrotz/peract_bimanual}{https://github.com/markusgrotz/peract\_bimanual}
    \item PerAct2: \href{https://github.com/markusgrotz/peract_bimanual}{https://github.com/markusgrotz/peract\_bimanual}
    \item DP-J: \href{https://github.com/real-stanford/diffusion_policy}{https://github.com/real-stanford/diffusion\_policy}
    \item DP3: \href{https://github.com/YanjieZe/3D-Diffusion-Policy}{https://github.com/YanjieZe/3D-Diffusion-Policy}
\end{itemize}

\begin{table*}[ht]
  \centering
  \caption{The experimental result on simulated tasks. We train the policy with the setting of different training demonstrations, \ie $[20, 100]$, to test its capability comprehensively. The best results are in \textbf{bold}. Each result is reported with three seeds on average.}
  \resizebox{0.95\linewidth}{!}{
    \begin{tabular}{l>{\raggedleft\arraybackslash}p{1.6cm}>{\raggedleft\arraybackslash}p{1.6cm}>{\raggedleft\arraybackslash}p{1.7cm}>{\raggedleft\arraybackslash}p{1.6cm}>{\raggedleft\arraybackslash}p{1.6cm}>{\raggedleft\arraybackslash}p{1.6cm}}
    \toprule[1pt]
          & \multicolumn{1}{r}{\makecell{Push\\Box}} & \multicolumn{1}{c}{\makecell{Lift\\Ball}} & \multicolumn{1}{r}{\makecell{Handover\\Item (easy)}} & \multicolumn{1}{r}{\makecell{Pick\\Laptop}} & \multicolumn{1}{r}{\makecell{Sweep\\Dustpan}} & \multicolumn{1}{c}{Overall} \\ [0.5ex]
    \midrule\midrule
    \multicolumn{1}{l}{\textit{\textbf{Demonstration Quantity}}}     &       &       &       &       &  \\
    \midrule
    KStar Diffuser (\textit{num demos=20}) & 79.3\textcolor[gray]{0.5}{\small{$\ \,\pm3.5$}} & 87.0\textcolor[gray]{0.5}{\small{$\ \,\pm2.7$}} & 23.7\textcolor[gray]{0.5}{\small{$\ \,\pm0.6$}} & 17.0\textcolor[gray]{0.5}{\small{$\ \,\pm2.0$}} & 83.0\textcolor[gray]{0.5}{\small{$\ \,\pm4.4$}} & 58.0\textcolor[gray]{0.5}{\small{$\ \,\pm1.4$}} \\
    KStar Diffuser (\textit{num demos=50}) & 81.0\textcolor[gray]{0.5}{\small{$\ \,\pm3.6$}} & 94.3\textcolor[gray]{0.5}{\small{$\ \,\pm2.5$}} & 24.3\textcolor[gray]{0.5}{\small{$\ \,\pm3.1$}} & 27.7\textcolor[gray]{0.5}{\small{$\ \,\pm3.2$}} & 85.0\textcolor[gray]{0.5}{\small{$\ \,\pm2.7$}} & 62.5\textcolor[gray]{0.5}{\small{$\ \,\pm1.4$}} \\
    KStar Diffuser (\textit{num demos=100}) & \textbf{83.0}\textcolor[gray]{0.5}{\small{$\ \,\pm1.7$}} & \textbf{98.7}\textcolor[gray]{0.5}{\small{$\ \,\pm1.5$}} & \textbf{27.0}\textcolor[gray]{0.5}{\small{$\ \,\pm1.7$}} & \textbf{43.7}\textcolor[gray]{0.5}{\small{$\ \,\pm4.5$}} & \textbf{89.0}\textcolor[gray]{0.5}{\small{$\ \,\pm5.2$}} & \textbf{68.2}\textcolor[gray]{0.5}{\small{$\ \,\pm2.1$}} \\

    \midrule\midrule
    \multicolumn{1}{l}{\textit{\textbf{Action Chunking Length}}}     &       &       &       &       &  \\
    \midrule
    KStar Diffuser (\textit{chunking=1}) & \textbf{92.0}\textcolor[gray]{0.5}{\small{$\ \,\pm1.7$}} & \textbf{98.7}\textcolor[gray]{0.5}{\small{$\ \,\pm0.6$}} & 23.7\textcolor[gray]{0.5}{\small{$\ \,\pm5.9$}} & 16.0\textcolor[gray]{0.5}{\small{$\ \,\pm5.2$}} & 12.3\textcolor[gray]{0.5}{\small{$\ \,\pm4.2$}} & 48.5\textcolor[gray]{0.5}{\small{$\ \,\pm0.6$}} \\

    KStar Diffuser (\textit{chunking=2}) & 83.0\textcolor[gray]{0.5}{\small{$\ \,\pm1.7$}} & \textbf{98.7}\textcolor[gray]{0.5}{\small{$\ \,\pm1.5$}} & \textbf{27.0}\textcolor[gray]{0.5}{\small{$\ \,\pm1.7$}} & \textbf{43.7}\textcolor[gray]{0.5}{\small{$\ \,\pm4.5$}} & 89.0\textcolor[gray]{0.5}{\small{$\ \,\pm5.2$}} & \textbf{68.2}\textcolor[gray]{0.5}{\small{$\ \,\pm2.1$}} \\

    KStar Diffuser (\textit{chunking=5}) & 81.3\textcolor[gray]{0.5}{\small{$\ \,\pm1.5$}} & 97.7\textcolor[gray]{0.5}{\small{$\ \,\pm2.5$}} & 1.7\textcolor[gray]{0.5}{\small{$\ \,\pm1.2$}} & 14.3\textcolor[gray]{0.5}{\small{$\ \,\pm3.2$}} & \textbf{99.3}\textcolor[gray]{0.5}{\small{$\ \,\pm1.2$}} & 58.9\textcolor[gray]{0.5}{\small{$\ \,\pm1.4$}} \\

    \midrule\midrule
    \multicolumn{1}{l}{\textit{\textbf{Historical Observation Length}}}     &       &       &       &       &  \\
    \midrule
    KStar Diffuser (\textit{history=0}) & 86.3\textcolor[gray]{0.5}{\small{$\ \,\pm3.5$}} & \textbf{99.7}\textcolor[gray]{0.5}{\small{$\ \,\pm0.6$}} & 9.7\textcolor[gray]{0.5}{\small{$\ \,\pm2.1$}} & 0.0\textcolor[gray]{0.5}{\small{$\ \,\pm0.0$}} & 0.0\textcolor[gray]{0.5}{\small{$\ \,\pm0.0$}} & 39.1\textcolor[gray]{0.5}{\small{$\ \,\pm1.1$}} \\
    KStar Diffuser (\textit{history=1}) & \textbf{87.7}\textcolor[gray]{0.5}{\small{$\ \,\pm4.0$}} & 28.3\textcolor[gray]{0.5}{\small{$\ \,\pm2.5$}} & 10.3\textcolor[gray]{0.5}{\small{$\ \,\pm7.4$}} & 36.3\textcolor[gray]{0.5}{\small{$\ \,\pm8.5$}} & \textbf{92.7}\textcolor[gray]{0.5}{\small{$\ \,\pm1.5$}} & 51.1\textcolor[gray]{0.5}{\small{$\ \,\pm1.5$}} \\
    KStar Diffuser (\textit{history=2}) & 83.0\textcolor[gray]{0.5}{\small{$\ \,\pm1.7$}} & 98.7\textcolor[gray]{0.5}{\small{$\ \,\pm1.5$}} & \textbf{27.0}\textcolor[gray]{0.5}{\small{$\ \,\pm1.7$}} & \textbf{43.7}\textcolor[gray]{0.5}{\small{$\ \,\pm4.5$}} & 89.0\textcolor[gray]{0.5}{\small{$\ \,\pm5.2$}} & \textbf{68.2}\textcolor[gray]{0.5}{\small{$\ \,\pm2.1$}} \\

    \midrule\midrule
    \multicolumn{1}{l}{\textit{\textbf{Coefficient $\lambda$}}}     &       &       &       &       &  \\
    \midrule
    KStar Diffuser (\textit{$\lambda$=0.1}) & \textbf{84.3}\textcolor[gray]{0.5}{\small{$\ \,\pm4.9$}} & \textbf{99.3}\textcolor[gray]{0.5}{\small{$\ \,\pm1.2$}} & 12.7\textcolor[gray]{0.5}{\small{$\ \,\pm2.3$}} & 19.7\textcolor[gray]{0.5}{\small{$\ \,\pm4.7$}} & 89.7\textcolor[gray]{0.5}{\small{$\ \,\pm2.5$}} & 61.1\textcolor[gray]{0.5}{\small{$\ \,\pm1.1$}} \\
    KStar Diffuser (\textit{$\lambda$=0.5}) & 83.7\textcolor[gray]{0.5}{\small{$\ \,\pm11.6$}} & 98.3\textcolor[gray]{0.5}{\small{$\ \,\pm1.5$}} & 3.3\textcolor[gray]{0.5}{\small{$\ \,\pm1.2$}} & 36.3\textcolor[gray]{0.5}{\small{$\ \,\pm6.8$}} & \textbf{93.7}\textcolor[gray]{0.5}{\small{$\ \,\pm3.8$}} & 63.1\textcolor[gray]{0.5}{\small{$\ \,\pm4.5$}} \\
    KStar Diffuser (\textit{$\lambda$=0.9}) & 83.0\textcolor[gray]{0.5}{\small{$\ \,\pm1.7$}} & 98.7\textcolor[gray]{0.5}{\small{$\ \,\pm1.5$}} & \textbf{27.0}\textcolor[gray]{0.5}{\small{$\ \,\pm1.7$}} & \textbf{43.7}\textcolor[gray]{0.5}{\small{$\ \,\pm4.5$}} & 89.0\textcolor[gray]{0.5}{\small{$\ \,\pm5.2$}} & \textbf{68.2}\textcolor[gray]{0.5}{\small{$\ \,\pm2.1$}} \\
    \bottomrule[1pt]
    \end{tabular}
    }%
  \label{tab:appendix_exp}%
\end{table*}%

\section{Extensive Ablation Studies}\label{appendix:sec_extensive_ablation}
To evaluate KStar Diffuser more comprehensively, we conduct following extensive experiments.

\vspace{-4mm}
\paragraph{(1) The Effects of Demonstration Quantity.} 
Given the critical role of demonstration quantity in imitation learning, we conduct an ablation study by training the KStar diffuser with 50 demonstrations, with results reported in Table~\ref{tab:appendix_exp}. 
Additionally, we provide a comparative analysis of policy performance across varying numbers of demonstrations (20, 50, and 100), as shown in Figure 5.
The results demonstrate \textbf{a clear positive correlation between demonstration quantity and policy performance}. 
With 20 demonstrations, the policy achieves basic task completion capabilities. 
Increasing to 50 demonstrations yields significant performance improvements, with success rates rising by around 4.5\% across the task suite. 
The upward trend continues as we scale to 100 demonstrations, indicating that the policy benefits from larger demonstration sets. 
These findings suggest that expanding the demonstration dataset consistently enhances the policy's ability to learn and generalize manipulation tasks.

\begin{figure}[ht]
    \centering
    \includegraphics[width=\linewidth]{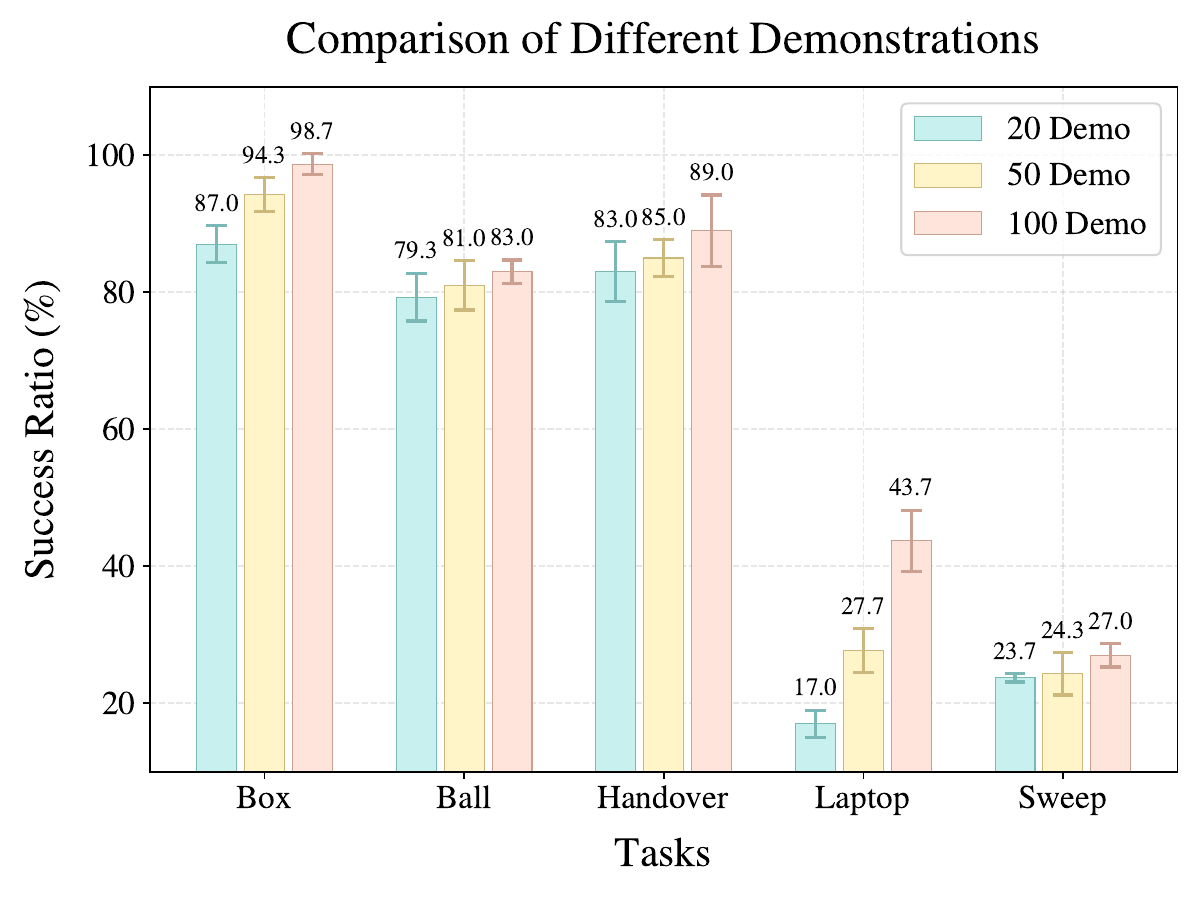}
    \setlength{\abovecaptionskip}{-10pt}
    \caption{The comparison of different number of demonstrations.}
    \vspace{-2.5mm}
    \label{fig:framwork}
\end{figure}

\begin{figure*}[ht]
    \centering
    \vspace{8pt}
    \includegraphics[width=\linewidth]{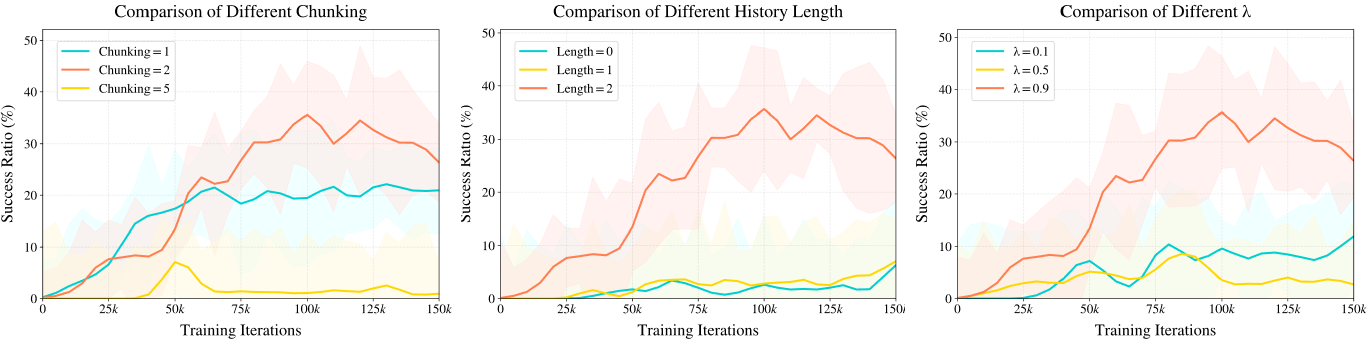}
    \caption{\textit{The left}: The result of different action chunking size. \textit{The middle}: The result of history lengths. \textit{The right}: The result of different coefficient $\lambda$.}
    \label{fig:appendix_exp}
    \vspace{-4mm}
\end{figure*}
\vspace{-4mm}
\paragraph{(2) The Effects of Action Chunking Size.}
As mentioned in previous work~\cite{chi2023diffusion, zhao2023learning}, action chunking prediction serves as an effective approach to address the multimodality challenges inherent in robotic manipulation tasks. 
To empirically verify this mechanism within our framework, we conducted an evaluation of the KStar Diffuser under various action chunking size configurations.
Our experimental setup explores three distinct action chunking strategies: 1) single-step prediction where only the next optimal pose is generated (action chunk = 1), 2) two-step prediction of consecutive optimal poses (action chunk = 2), and 3) five-step prediction of sequential optimal poses (action chunk = 5).

The results presented in Table~\ref{tab:appendix_exp} reveal an \textbf{interesting trade-off between prediction horizon and model performance}. 
While single-step prediction (action chunk = 1) provides basic capabilities, it usually meets the multimodal problem.
The two-step prediction strategy (action chunk = 2) emerges as the optimal configuration, demonstrating superior success rates and motion quality across all tasks. 
Notably, attempting to predict longer sequences (action chunk = 5) leads to decreased performance, with success rates dropping by around 10\% compared to the two-step configuration. 
This performance degradation suggests that when predicting next best pose, extended prediction horizons introduce excessive complexity into the learning problem, making it challenging for the policy to capture and generate accurate action sequences. 
As shown in Figure~\ref{fig:appendix_exp}, we present the variation in the success ratio for the \texttt{handover\_item\_easy} task, under different action chunking size configurations as the number of training steps increases.

\begin{figure*}[htb]
    \centering
    \includegraphics[width=\linewidth]{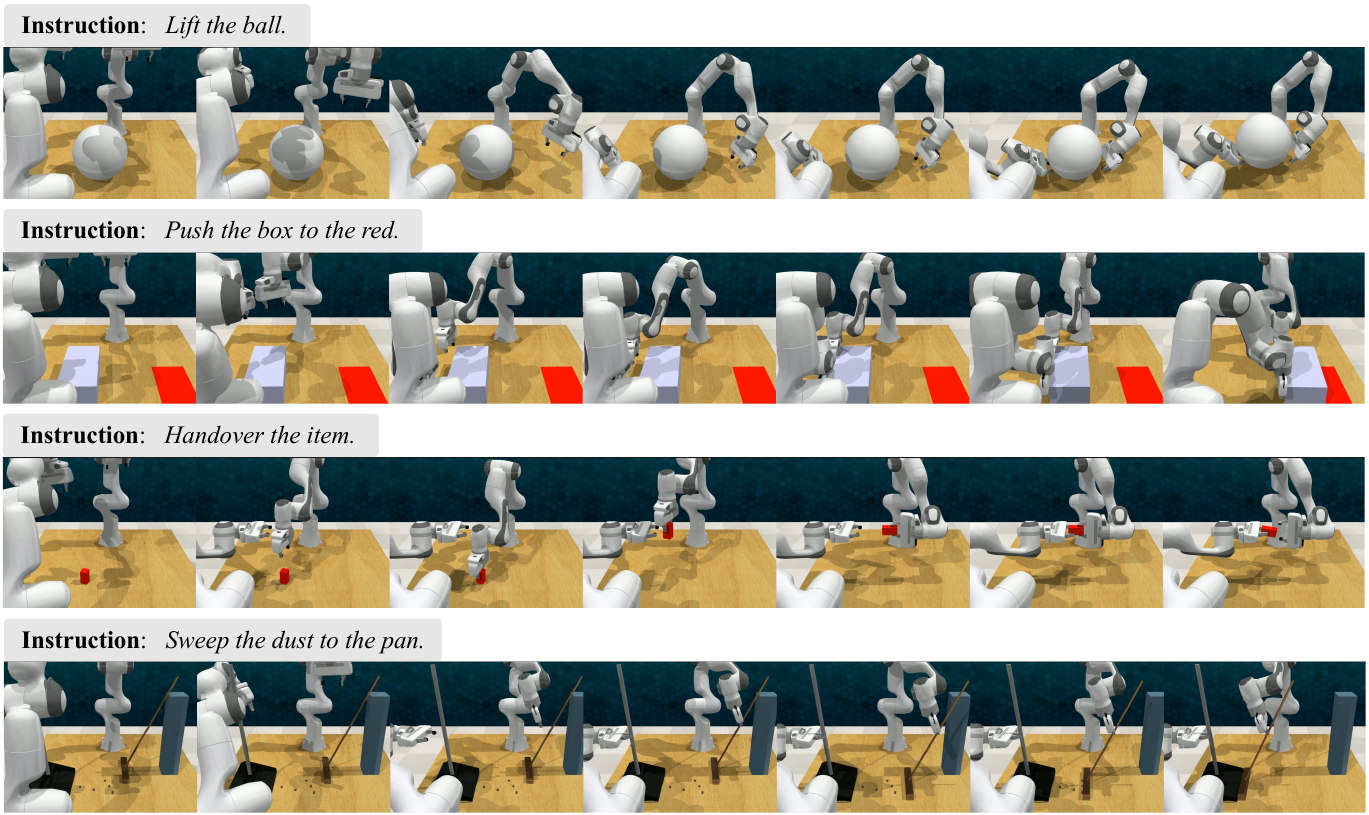}
    \caption{The visualization of simulated tasks, including \texttt{push\_box}, \texttt{lift\_ball}, \texttt{handover\_item\_easy}, \texttt{sweep\_dustpan}.}
    \label{fig:appendix_visualization}
\end{figure*}
\vspace{-3mm}
\paragraph{(3) The Effects of Historical Observation Length.}
During our experiments, we found that the historical information plays an important role in action prediction quality.
We conduct an ablation study across three history lengths: 0 (current observation only), 1, and 2 steps.
The results are shown in Table~\ref{tab:appendix_exp}.
With no historical information (0-step), the policy fails to learn effective policies, as it cannot capture the temporal dependencies crucial for manipulation tasks.
Adding one historical step enables basic learning capabilities, with the model achieving preliminary success in simpler tasks.
Further extending to two historical steps yields optimal performance, showing an approximately 17\% improvement over the single-step configuration and demonstrating enhanced stability across all tasks.
While longer history lengths might provide more temporal context, they risk introducing redundant information that could potentially obscure relevant features, as observed in our preliminary experiments.
These findings suggest that while temporal context is essential for understanding the current state and predicting actions, \textbf{an appropriate history window}, \eg, 2 steps, provides an optimal balance between capturing necessary temporal dependencies and maintaining computational efficiency.
As shown in Figure~\ref{fig:appendix_exp}, we present the variation in the success ratio for the \texttt{handover\_item\_easy} task, under different history length configurations as the number of training steps increases.

\vspace{-4mm}
\paragraph{(4) The Effects of Coefficient $\lambda$.}
In policy learning, a kinematic regularization term is incorporated into the next best pose learning objective to balance task effectiveness and motion constrains. 
The regularization strength is controlled by the coefficient $\lambda$, which determines the extent of kinematic constraints imposed on the learning process.
The choice of $\lambda$ significantly influences policy performance. Larger values of $\lambda$ (approaching 1.0) correspond to weaker kinematic constraints, enabling more flexible motion patterns. Conversely, smaller values of $\lambda$ (e.g., 0.1) impose stricter kinematic constraints, yielding more conservative policies that emphasize motion smoothness over task efficiency.
Empirical results, as shown in Table~\ref{tab:appendix_exp}, demonstrate that policy performance reaches its optimum at $\lambda = 0.9$. For smaller values of $\lambda$, the learned policies exhibit overly cautious behavior, manifesting in trajectories that prioritize smoothness at the expense of task efficiency and optimal path planning. Conversely, as $\lambda$ approaches 1.0, the diminished kinematic constraints result in less regulated motion patterns, potentially compromising trajectory naturalness and precision.
These findings indicate that $\lambda = 0.9$ achieves a better trade-off between preserving natural motion characteristics and ensuring efficient task execution. 
This configuration effectively minimizes the adverse effects of both excessive motion constraints and insufficient regulation, ultimately yielding superior performance across our task suite.
As shown in Figure~\ref{fig:appendix_exp}, we present the variation in the success ratio for the \texttt{handover\_item\_easy} task, under different coefficient $\lambda$ configurations as the number of training steps increases.

\section{Qualitative Analysis}\label{appendix:qualitative_analysis}
We show more qualitative result in Figure~\ref{fig:appendix_visualization}.
Through the novel approach of encoding robotic arm structural information as graph representations and explicitly incorporating kinematic constraints, our model demonstrates exceptional performance in motion symmetry, synchronization, and coordination across dual-arm manipulation tasks. 
In the \texttt{lift\_ball} experiment, the model achieves precise bilateral symmetry in spatial positioning through learned structured representations. 
The dual arms maintain stable symmetric configurations while preventing object instability through synchronized force application patterns, highlighting the efficacy of our structure-aware control paradigm.
Furthermore, in the \texttt{push\_box} task, the model exhibits remarkable geometric symmetry in motion planning and execution. 
By leveraging embedded kinematic information, the robotic arms consistently maintain equidistant positioning relative to the target object's center of mass while executing synchronized trajectories along parallel paths. 
This precise symmetrical control not only ensures operational stability during object manipulation but also establishes a robust framework for dual-arm cooperative control in complex manipulation scenarios.